%% file: trillion.tex
\documentclass[11pt, a4paper, logo, copyright, nonumbering]{trillion}

\usepackage{silence}
\usepackage[authoryear, sort&compress, round]{natbib}
\usepackage{dblfloatfix}
\usepackage{ulem}
\usepackage{caption}
\usepackage{xspace}
\usepackage{pifont} %
\usepackage{booktabs}
\usepackage{multirow}
\usepackage{tcolorbox}
\usepackage{xltabular}
\usepackage{longtable}
\usepackage{hyperref}
\usepackage{amsfonts}
\usepackage{amsmath}
\usepackage{amssymb}
\usepackage{lineno}
\usepackage{multirow}
\usepackage{adjustbox}
\usepackage{graphicx}
\usepackage[bottom]{footmisc}

\usepackage{CJKutf8}
\usepackage{subfigure}
\usepackage{setspace}

\usepackage{dsfont}
\usepackage{array} %
\usepackage{tabularx} %
\usepackage{subfigure} 
\usepackage{xcolor} 

\usepackage{pgfplots}
\pgfplotsset{compat=1.18}
\usetikzlibrary{shapes.geometric}
\usepackage{lipsum}  
\usepackage{multicol} 
\usepackage{xcolor}
\definecolor{mypurple}{HTML}{3C2C63}

\usepackage[T1]{fontenc}
\usepackage{amsmath,amsfonts,amssymb,amsthm,mathtools,bm}
\usepackage{booktabs}
\usepackage{nicefrac}
\usepackage[tracking=smallcaps]{microtype}
\usepackage{xcolor}
\usepackage{enumitem}
\usepackage{multirow}
\usepackage{array}
\usepackage{color}
\usepackage[table]{xcolor} 
\usepackage{wrapfig}
\usepackage{multicol}
\usepackage{caption}
\usepackage{diagbox}
\usepackage{pifont}
\usepackage{pdfpages}
\usepackage{arydshln}
\usepackage{makecell}
\usepackage{graphicx}
\usepackage{adjustbox}
\usepackage{colortbl}
\usepackage{siunitx}
\usepackage{pifont}
\usepackage{seqsplit}
\usepackage{fancyvrb}
\tcbuselibrary{breakable}
\usepackage{fvextra}

\definecolor{blue(pigment)}{rgb}{0.2, 0.2, 0.6}
\definecolor{aliceblue}{rgb}{0.94, 0.97, 1.0}
\definecolor{lightgray}{rgb}{0.88, 0.88, 0.88}
\definecolor{piggypink}{rgb}{0.95, 0.9, 0.96}
\definecolor{mistyrose}{rgb}{1.0, 0.89, 0.88}
\definecolor{deeppurple}{rgb}{0.42, 0.13, 0.42}
\definecolor{lightgray}{rgb}{0.83, 0.83, 0.83}
\definecolor{pastelgray}{rgb}{0.81, 0.81, 0.77}
\definecolor{grey}{rgb}{0.5,0.5,0.5}
\definecolor{formatcolor}{rgb}{0.01, 0.31, 0.59}

\hypersetup{
  colorlinks=true,
  linkcolor=formatcolor,
  citecolor=formatcolor
}
\usepackage[capitalise,noabbrev,nameinlink]{cleveref}

\crefformat{figure}{Figure~#2{\color{formatcolor}#1}#3}
\crefformat{table}{Table~#2{\color{formatcolor}#1}#3}
\crefformat{equation}{Equation~#2{\color{formatcolor}#1}#3}
\crefformat{section}{Section~#2{\color{formatcolor}#1}#3}
\crefformat{appendix}{Appendix~#2{\color{formatcolor}#1}#3}

\newcommand{\huggingface}{\raisebox{-1.5pt}{\includegraphics[height=1.05em]{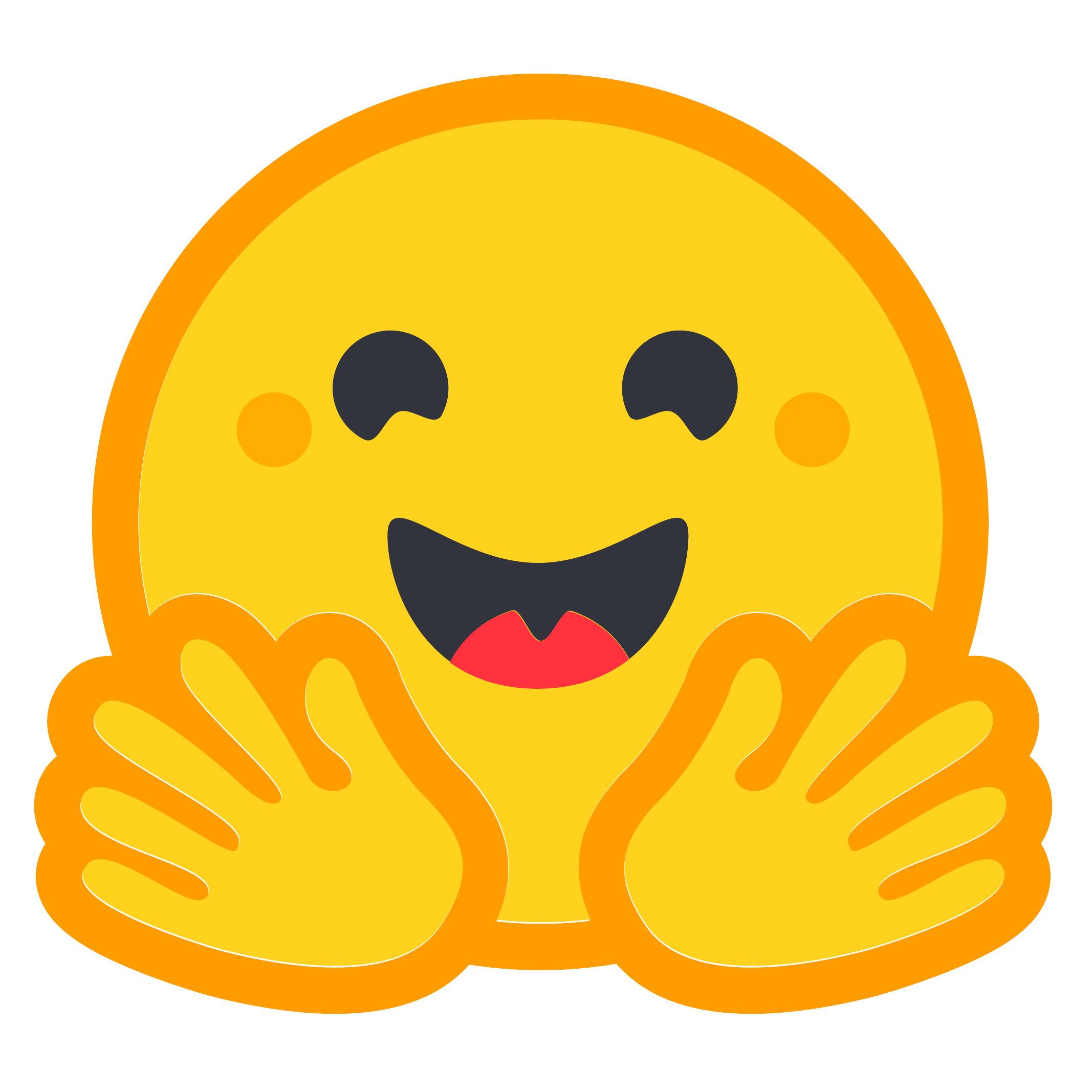}}\xspace}

\makeatletter
\def\@BTrule[#1]{%
  \ifx\longtable\undefined
    \let\@BTswitch\@BTnormal
  \else\ifx\hline\LT@hline
    \nobreak
    \let\@BTswitch\@BLTrule
  \else
     \let\@BTswitch\@BTnormal
  \fi\fi
  \global\@thisrulewidth=#1\relax
  \ifnum\@thisruleclass=\tw@\vskip\@aboverulesep\else
  \ifnum\@lastruleclass=\z@\vskip\@aboverulesep\else
  \ifnum\@lastruleclass=\@ne\vskip\doublerulesep\fi\fi\fi
  \@BTswitch}
\makeatother

\addto\extrasenglish{
    \def\sectionautorefname{Section}%

}

 {\begin{list}{}%
         {\setlength{\leftmargin}{#1}}%
         \item[]%
 }
 {\end{list}}
 
\reportnumber{} %

\renewcommand{\today}{}

\title{\centering Diffusion Drafts, AR Verifies: Accelerating Document OCR with Self-Speculative Decoding}

\renewcommand\Affilfont{\normalfont\fontsize{10}{12}\selectfont\centering}
\author[1,2]{\textcolor{black}{Dohyun~Kim}}
\author[1]{\textcolor{black}{Sungjun~Han}}
\author[1]{\textcolor{black}{Hyungguk~Kim}}
\author[1]{\textcolor{black}{Yusik~Kim}}
\author[1]{Jamin~Shin}
\author[2]{\textcolor{black}{Paul~Hongsuck~Seo}}
\author[1,3]{\textcolor{black}{Hongjoon~Ahn}}
\affil[1]{\textbf{Trillion Labs}}
\affil[2]{\textbf{Korea University}}
\affil[3]{\textbf{Seoul National University}}

\date{}

\input{macros.tex}

\input{sections/00_abstract}

\begin{document}
\begin{CJK*}{UTF8}{mj}

\maketitle

\input{sections/01_intro}

\input{sections/02_preliminary}

\input{sections/04_method}

\input{sections/05_experiment}

\input{sections/06_related_work}

\input{sections/07_conclusion}

\newpage
\bibliography{bib}

\newpage
\appendix
\renewcommand{\sectionautorefname}{Appendix}
\renewcommand{\subsectionautorefname}{Appendix} %
\crefalias{section}{appendix}
\crefalias{subsection}{appendix}

\input{sections/appendix}

\end{CJK*}
\end{document}

%% file: macros.tex
\definecolor{provgray}{gray}{0.55}

\providecommand{\system}{}\renewcommand{\system}{\textsc{GravityOCR}\xspace} %
\newcommand{\ours}{GravityOCR\xspace}
\newcommand{\glmocr}{GLM-OCR\xspace}
\newcommand{\odb}{OmniDocBench\xspace}

%% file: sections/00_abstract.tex
\begin{abstract}

Autoregressive OCR vision--language models accurately convert document images into text and structured markup, but require one sequential decoding step per output token, limiting inference speed. Unlike open-ended text generation, OCR outputs are strongly grounded in the input image, making diffusion-based parallel generation promising. However, when several tokens are predicted in one diffusion step, each is predicted before the others are known. Committing them directly can therefore introduce errors.
We therefore introduce \system{}, a parameter-shared AR--block-diffusion model jointly trained for parallel drafting and causal AR verification.
Verifying drafts before commitment lets the model commit multiple output tokens per round without a separate drafting network.
The causal AR path also enables GRPO with sequence- and structure-level OCR rewards, avoiding diffusion-trajectory likelihood estimation while updating the shared drafter parameters.
On \odb{} v1.6, AR-path GRPO improves the Overall score from 94.92 to 95.16
without reducing diffusion drafting efficiency, while the final model remains
close to the original \glmocr{} score of 95.48.
In an SGLang serving deployment, \system{} commits an average of 9.7 output tokens per forward pass and achieves a \textbf{3.94$\times$ decode-only speedup} on region crops and a \textbf{1.32$\times$ end-to-end page-processing speedup} over AR decoding.
\par\medskip
{\centering
\href{https://github.com/trillion-labs/GravityOCR}{\textcolor{formatcolor}{\raisebox{-1.5pt}{\includegraphics[trim=70 70 70 70,clip,height=1.1em]{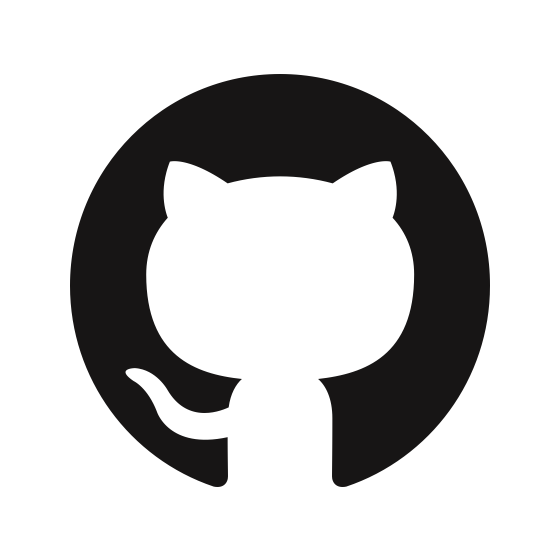}}\enspace\textbf{Code \& Documentation}}}
\hspace{1.8em}
\href{https://huggingface.co/trillionlabs/GravityOCR}{\textcolor{formatcolor}{\huggingface\enspace\textbf{Model Weights}}}
\par}

\end{abstract}

%% file: sections/01_intro.tex
\section{Introduction}
\label{sec:intro}

\input{figures/teaser}

Generative OCR vision--language models have emerged as a leading approach to
document understanding, converting document images into serialized sequences
of plain text and structured markup such as HTML and LaTeX.
Most of these models generate their outputs autoregressively, requiring one
sequential decoding step for each output token.
This sequential dependency directly affects latency and throughput when OCR
models are deployed at scale.
Recent OCR systems have therefore explored faster generation through
multi-token prediction and speculative decoding
\citep{leviathan2023fast,chen2023accelerating,duan2026glmocr,li2026hunyuanocr15}.
In parallel, diffusion-based OCR models have begun to generate multiple output
tokens simultaneously
\citep{dong2026mineru,man2026dodo}.

OCR is particularly amenable to parallel generation because many output
positions are directly constrained by the input image, unlike in open-ended
text generation
\citep{man2026dodo,dong2026mineru}.
Masked diffusion provides a natural mechanism for exploiting this property:
multiple output positions can be recovered together rather than generated
strictly one at a time
\citep{austin2021d3pm,sahoo2024mdlm}.
This makes diffusion-based decoding a promising direction for reducing the
inference cost of OCR.

\begin{figure}[t!]
\centering
\includegraphics[width=0.98\linewidth]
{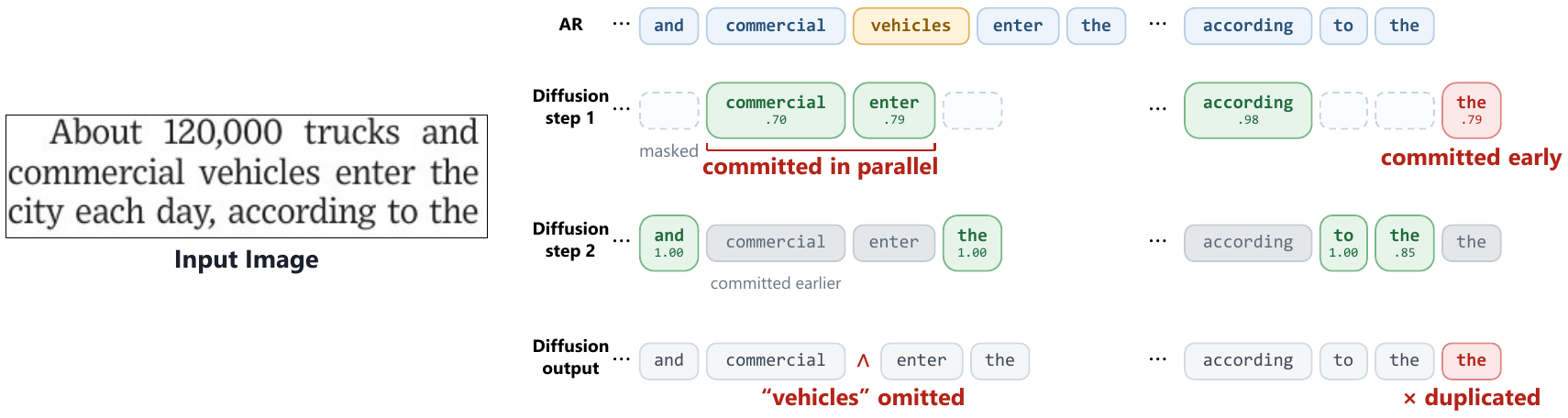}
\caption{\textbf{Failure case of confidence-based parallel commitment.} At $\tau_c{=}0.7$, \emph{commercial} and \emph{enter} are committed together while the intervening position remains unresolved, ultimately omitting \emph{vehicles}. Similarly, the final \emph{the} is committed before the two positions to its left are resolved, producing the repeated span \emph{to the the}. The causal AR output from the same checkpoint is shown for comparison.}
\label{fig:failure}
\end{figure}

Visual grounding, however, does not eliminate sequential dependencies among
OCR output tokens.
Text must preserve its reading order, while tags and delimiters in tables and
formulas must remain structurally consistent.
Under causal AR decoding, each token is predicted after the preceding tokens
have become available.
In a parallel diffusion step, by contrast, positions predicted together
cannot condition on the tokens simultaneously selected at the other
positions.
Directly committing these predictions can therefore produce tokens that are
individually plausible under the partially resolved block but inconsistent in
the completed output.
\Cref{fig:failure} illustrates this failure mode under confidence-based
parallel decoding
\citep{wu2025fastdllm}.

Motivated by this failure mode, we introduce \system{}, a parameter-shared
AR--diffusion model inspired by recent AR--diffusion hybrids and
diffusion-based drafting methods
\citep{fu2026nemotron,liu2025tidar,wu2026fastdvlm,chen2026dflash}.
\system{} uses block diffusion
\citep{arriola2025block} to propose an entire token block in one forward pass
and causal AR decoding to verify the proposal before commitment.
The causal path commits only the longest draft prefix consistent with its
token-by-token predictions.
With top-1 verification, the resulting self-speculative decoder reproduces the causal AR greedy output in exact arithmetic (\cref{app:decoding:equality} measures the agreement under bf16 serving kernels).
Successful drafts allow decoding to advance by multiple tokens, whereas a
draft--verifier disagreement limits the accepted draft length and attainable
acceleration.
Because both prediction paths share the same parameters, the method requires
neither a separate drafting network nor an auxiliary prediction head.

We instantiate \system{} by adapting \glmocr{} \citep{duan2026glmocr}.
The causal AR verifier determines the final output, so we apply GRPO
through this path using sequence- and structure-level OCR rewards.
This updates the shared parameters without estimating likelihoods over
diffusion trajectories.
On \odb{} v1.6, our final model retains 99.7\% of the original
\glmocr{} model's Overall score (95.16 vs.\ 95.48), while
self-speculative decoding substantially reduces the number of sequential decoding steps (\cref{fig:teaser}).

Our contributions are summarized as follows:
\begin{itemize}
    \item We show that a pretrained causal AR OCR model can be adapted into a
    parameter-shared block-diffusion drafter and causal AR verifier, enabling
    faster decoding with minimal accuracy degradation and without requiring a
    separate drafting network.

    \item We implement full-backbone self-speculative decoding in SGLang
    \citep{zheng2024sglang}, achieving 9.7 tokens per forward (TPF), where
    both diffusion drafting and causal verification forwards are counted.
    This yields a \(3.94\times\) speedup in token generation and a
    \(1.32\times\) speedup in end-to-end page processing over causal AR
    decoding. 

    \item We compare direct block-diffusion and self-speculative decoding on
    OCR, characterizing their accuracy--parallelism trade-off. We further
    show that GRPO applied only through the causal AR path improves the
    OmniDocBench Overall score from 94.92 to 95.16 while preserving the
    efficiency of the shared diffusion drafter.

\end{itemize}

%% file: figures/teaser.tex
\begin{figure}[t!]
\centering
\definecolor{vizgray}{HTML}{45443F}
\definecolor{vizblue}{HTML}{2A78D6}
\definecolor{vizorange}{HTML}{EB6834}
\definecolor{vizgrid}{HTML}{E4E4E0}
\definecolor{vizink}{HTML}{52514E}
\begin{tikzpicture}
\begin{axis}[
  width=0.80\linewidth, height=5.6cm, scale only axis,
  xmin=0, xmax=0.88, ymin=89.2, ymax=96.6,
  xlabel={single-stream speed on one H100 (pages/s)\,$\uparrow$},
  ylabel={\odb{} Overall\,$\uparrow$},
  xtick={0,0.2,0.4,0.6,0.8}, ytick={90,92,94,96},
  axis x line*=bottom, axis y line*=left,
  tick align=outside,
  every tick label/.append style={font=\scriptsize, color=vizink},
  label style={font=\small, color=vizink},
  axis line style={vizink, line width=0.5pt},
  ymajorgrids, major grid style={vizgrid, line width=0.35pt},
  clip=false,
]
\addplot[only marks, mark=*, mark size=2.6pt,
         mark options={fill=vizgray, draw=white, line width=0.7pt}] coordinates {
  (0.022,91.42) (0.389,94.86) (0.059,89.87) (0.116,91.14)
  (0.399,95.57) (0.407,93.18)};
\addplot[only marks, mark=*, mark size=2.6pt,
         mark options={fill=vizgray, draw=white, line width=0.7pt}] coordinates {(0.579,95.52)};
\addplot[only marks, mark=*, mark size=2.8pt,
         mark options={fill=vizblue, draw=white, line width=0.7pt}] coordinates {(0.571,95.48)};
\node[star, star points=5, star point ratio=2.35, minimum size=13pt,
      inner sep=0pt, fill=vizorange, draw=white, line width=0.9pt]
  at (axis cs:0.730,95.16) {};
\node[font=\scriptsize, color=vizgray, anchor=west,  xshift=4pt] at (axis cs:0.022,91.42) {DeepSeek-OCR-2};
\node[font=\scriptsize, color=vizgray, anchor=east,  xshift=-4pt] at (axis cs:0.389,94.86) {PaddleOCR-VL-1.5};
\node[font=\scriptsize, color=vizgray, anchor=west,  xshift=4pt] at (axis cs:0.059,89.87) {MinerU-Diffusion};
\node[font=\scriptsize, color=vizgray, anchor=west,  xshift=4pt] at (axis cs:0.116,91.14) {dots.ocr};
\node[font=\scriptsize, color=vizgray, anchor=south east, xshift=3pt, yshift=4pt] at (axis cs:0.399,95.57) {MinerU2.5-Pro};
\node[font=\scriptsize, color=vizgray, anchor=west,  xshift=4pt] at (axis cs:0.407,93.18) {MinerU2.5};
\node[font=\scriptsize, color=vizgray, anchor=north, xshift=-8pt, yshift=-6pt] at (axis cs:0.579,95.52) {HunyuanOCR-1.5};
\node[font=\scriptsize, color=vizblue, anchor=south, yshift=6pt] at (axis cs:0.571,95.48) {GLM-OCR};
\node[font=\small\bfseries, color=vizorange, anchor=south, yshift=8pt] at (axis cs:0.730,95.16) {\ours};
\end{axis}
\end{tikzpicture}
\caption{\textbf{Accuracy--speed landscape of document OCR systems.} \odb{} v1.6 Overall vs.\ single-stream page rate, with the same evaluation set used across systems for each metric. \ours{} is the fastest system measured while maintaining top-tier accuracy.}
\label{fig:teaser}
\end{figure}

%% file: sections/02_preliminary.tex
\section{Preliminaries}
\label{sec:prelim}

\paragraph{Task Formulation.}
Given an image
\(I\) and a task prompt \(c\), let
\(\mathbf{y}=(y_1,\ldots,y_N)\) denote the corresponding target token
sequence, which may encode plain text, an HTML table, or a LaTeX formula.
Let \(p_{\theta}^{\mathrm{AR}}\) denote the conditional distribution induced
by the decoder under causal attention.
The autoregressive likelihood of \(\mathbf{y}\) is factorized as
\begin{equation}
p_{\theta}^{\mathrm{AR}}(\mathbf{y} \mid I,c)
=
\prod_{i=1}^{N}
p_{\theta}^{\mathrm{AR}}
\left(y_i \mid I,c,\mathbf{y}_{<i}\right),
\end{equation}
with the negative log-likelihood objective
\begin{equation}
\mathcal{L}_{\mathrm{AR}}
=
-\sum_{i=1}^{N}
\log p_{\theta}^{\mathrm{AR}}
\left(y_i \mid I,c,\mathbf{y}_{<i}\right).
\label{eq:ar}
\end{equation}
At inference, each token is generated only after its preceding tokens become
available, so an output of length \(N\) requires \(N\) sequential decoding
steps.

\paragraph{Block Diffusion.}

Rather than generating one token at a time, masked diffusion language models
iteratively reconstruct multiple masked positions at each denoising step
\citep{austin2021d3pm,nie2025llada,ye2025dream}.
Block diffusion partitions the output sequence \(\mathbf{y}\) into contiguous
blocks \(\mathbf{y}^{(b)}\), each containing at most \(B\) tokens
\citep{arriola2025block,wu2025fastdllmv2}.
Here, \(b\) indexes the blocks in left-to-right order.
Generation is causal across blocks: the current block
\(\mathbf{y}^{(b)}\) conditions on the completed prefix
\(\mathbf{y}^{(<b)}
=[\mathbf{y}^{(1)},\ldots,\mathbf{y}^{(b-1)}]\).
Within the current block, attention is bidirectional.
Let \(p_{\theta}^{\mathrm{diff}}\) denote the conditional distribution induced
by the decoder under this block-causal attention pattern.
For a sampled noise level \(t\), let \(\mathbf{y}_t^{(b)}\) denote the corrupted
block obtained by replacing the positions in
\(\mathcal{M}_t^{(b)}\) with a dedicated mask token, denoted by
\(\mathtt{[M]}\).
The corresponding denoising objective is
\begin{equation}
\mathcal{L}_{\mathrm{diff}}
=
-\mathbb{E}_{b,t,\mathcal{M}_t^{(b)}}
\left[
w(t)
\sum_{j\in\mathcal{M}_t^{(b)}}
\log
p_{\theta}^{\mathrm{diff}}
\left(
y_j^{(b)}
\mid
I,c,\mathbf{y}^{(<b)},\mathbf{y}_t^{(b)}
\right)
\right],
\label{eq:diff}
\end{equation}
where \(j\) indexes token positions within block \(b\), and \(w(t)\) denotes
the timestep-dependent loss weight.

At inference, the current block is initialized with \(B\) mask tokens.
At each denoising step, the model predicts all remaining masked positions in
parallel, after which a selection rule---such as random selection,
confidence-based top-\(k\) selection, or confidence thresholding---determines
which predictions are unmasked and retained
\citep{sahoo2024mdlm,nie2025llada,wu2025fastdllm}.
The updated block is then used as input for the next denoising step.
Once the block is complete, it is appended to the prefix.
Because completed prefix blocks remain unchanged, their KV states can be
cached and reused when decoding subsequent blocks
\citep{arriola2025block,wu2025fastdllm}.

\paragraph{Speculative Decoding.}
Speculative decoding accelerates AR generation by allowing a drafter to
propose multiple future tokens, which the target AR model evaluates in
parallel
\citep{leviathan2023fast,chen2023accelerating}.
Under greedy decoding, consecutive draft tokens are accepted while they match
the target model's predictions.
Under stochastic decoding, the standard acceptance--rejection rule preserves
the target distribution.
Our shared-model realization is described in
\cref{sec:method:decode}.

%% file: sections/04_method.tex
\section{Method}
\label{sec:method}

Our goal is to reduce the decoding cost of OCR while retaining the
recognition accuracy of causal AR decoding.
Motivated by the failure mode analyzed in \cref{fig:failure}, we separate
parallel drafting from token commitment: a block-diffusion path proposes
multiple tokens in parallel, while a causal AR path verifies the proposals
before they are committed.
\Cref{sec:method:arloss,sec:method:decode,sec:method:rl} describe joint
AR--diffusion adaptation, self-speculative decoding, and reinforcement
learning through the causal AR path with sequence- and structure-level OCR
rewards, respectively.

\subsection{Learning a Shared Block-Diffusion Drafter and AR Verifier}
\label{sec:method:arloss}

\begin{figure}[t!]
    \centering
    \includegraphics[width=0.98\linewidth]{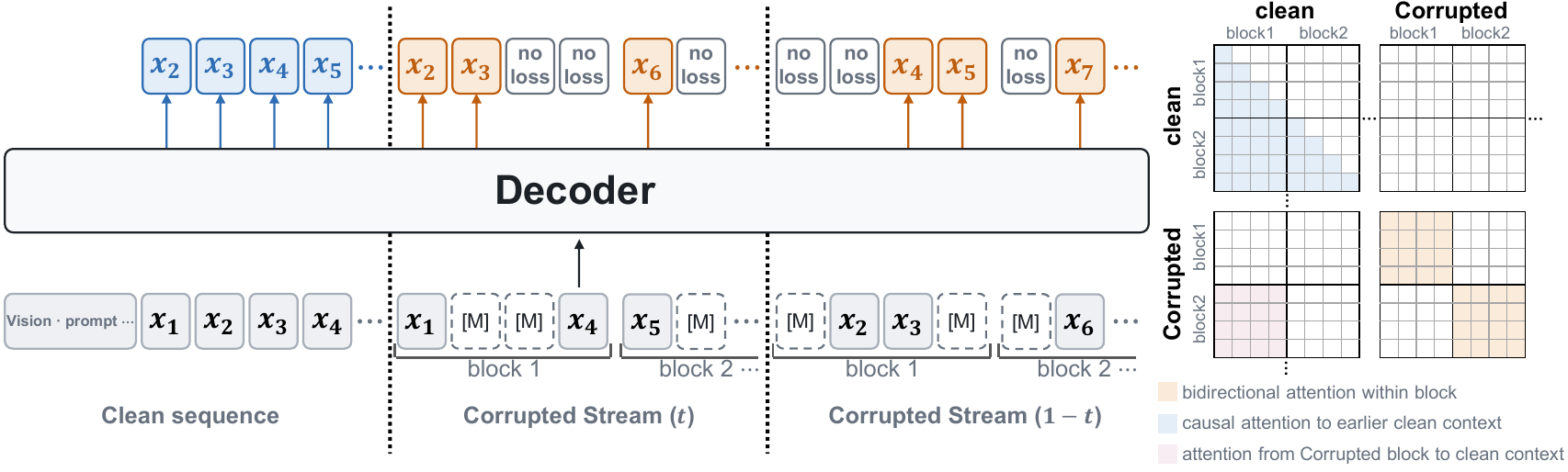}
    \caption{\textbf{Joint AR--diffusion training in a single forward pass.}
The clean response stream provides causal next-token supervision, while two
complementary corrupted streams with masking ratios \(t\) and \(1-t\)
provide denoising supervision over disjoint masked targets.
All streams share the visual and prompt context, decoder, and LM head.
The right panel contrasts token-level causal attention in the clean stream
with causal cross-block and bidirectional within-block attention in the
corrupted streams.}
\vspace{-1em}
    \label{fig:training}
\end{figure}

\paragraph{AR-to-Diffusion Conversion.}

We adapt a pretrained AR OCR model into a parameter-shared model that supports
both causal AR and block-diffusion prediction.
The two modes share the vision encoder, language decoder, and LM head; the
only architectural addition is a learned embedding for the dedicated mask
token, denoted by \(\mathtt{[M]}\) in our figures.
We introduce neither a separate drafter network nor an auxiliary prediction
head.

\paragraph{Joint AR--Diffusion Training.}
We jointly train the shared model for self-speculative decoding. Under block diffusion, the tokens decoded in the same step are predicted from the same partially masked block and cannot condition on one another's values, so committing them directly can leave the completed output inconsistent. We therefore also train the model on next-token prediction over the clean sequence, preserving the causal AR path that verifies each draft and rejects inconsistent proposals before they are committed.
\Cref{fig:training} provides an overview of our training procedure.
At each training step, the shared model processes one clean response stream alongside two complementary corrupted response streams in a single forward pass.
The clean stream provides causal next-token supervision for the AR verifier, whereas the corrupted streams provide denoising supervision for the block-diffusion drafter.
All three response streams are conditioned on the same image \(I\) and task prompt \(c\), both of which remain uncorrupted.

For each block \(\mathbf{y}^{(b)}\), we sample a noise level
\(t\sim\mathcal{U}(0,1)\) and mask each response token in the block
independently with probability \(t\), giving a set of masked indices
\(\mathcal{M}_t^{(b)}\).
The first corrupted stream replaces the tokens indexed by
\(\mathcal{M}_t^{(b)}\) with the mask token, whereas the second corrupted
stream masks the complementary positions of the same block.
The two streams therefore have expected masking ratios \(t\) and \(1-t\),
respectively.
Because the two masks are complementary, every response token, including the
\(B\) end-of-sequence tokens appended to each response (\cref{app:training}),
is masked and receives denoising supervision in exactly one of the two
streams, while image and prompt tokens are never masked or supervised.
This complementary masking scheme follows Fast-dLLM v2
\citep{wu2025fastdllmv2}.

The clean stream uses causal attention, so each target token \(y_i\) is predicted using only the image \(I\), task prompt \(c\), and preceding tokens \(\mathbf{y}_{<i}\).
Within each corrupted stream, tokens in block \(b\) can attend to all positions in the same corrupted block.
They can also attend to the preceding clean response blocks \(\mathbf{y}^{(<b)}\), but not to the current or future clean blocks.
Both clean and corrupted streams predict \(y_i\) from the logit at position \(i-1\). In corrupted streams, denoising supervision applies only to masked targets.

\paragraph{Training Objective.}
The losses from the clean and corrupted streams are combined as
\begin{equation}
\mathcal{L} %
=
\mathcal{L}_{\mathrm{AR}}
+
\lambda\mathcal{L}_{\mathrm{diff}},
\end{equation}
where \(\lambda\) controls the relative weight of the diffusion objective.
In implementation, we average the AR loss in \cref{eq:ar} over supervised
clean-stream tokens and the diffusion loss in \cref{eq:diff} over masked
targets pooled across both corrupted streams, using \(w(t)=1\).
We normalize the combined loss by \(1+\lambda\) and use \(\lambda=1\), giving
each objective weight 0.5, except in the no-AR-loss ablation in
\cref{tab:arloss_main}.

\subsection{Self-Speculative Decoding}
\label{sec:method:decode}

\begin{figure}[t!]
    \centering
    \includegraphics[width=0.98\linewidth]{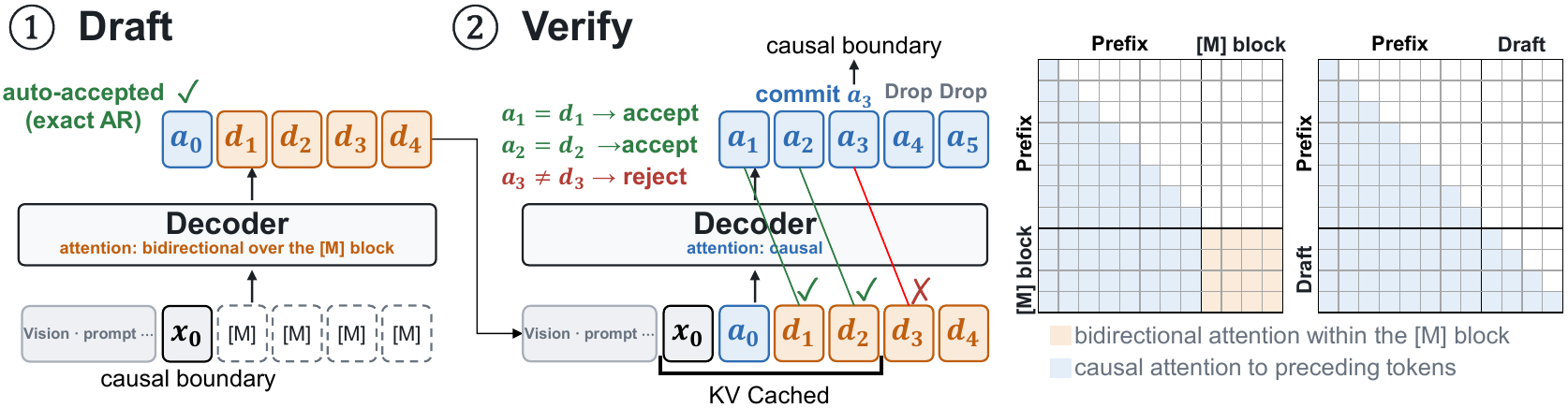}
    \caption{\textbf{Self-speculative decoding.} Given a causal boundary token \(x_0\) followed by \(B\) mask tokens, the shared decoder produces in a single forward pass one causal token \(a_0\) that is accepted by construction and \(B\) parallel draft tokens \(d_{1:B}\), attending bidirectionally within the mask block and causally to the cached prefix; the draft is passed on without iterative unmasking or confidence-based selection. A causal verification pass over \(a_0, d_{1:B}\) produces \(a_{1:B+1}\) and commits the longest AR-consistent draft prefix and the next AR token while simultaneously constructing the KV cache for the accepted prefix, so no additional cache-construction pass is needed; the states of rejected tokens are discarded. The final AR token becomes the causal boundary for the next round and obtains its KV state there.}
    \label{fig:inference}
\end{figure}

In direct block-diffusion decoding, token commitment is based on the
diffusion model's confidence.
An incorrect token can therefore be committed before the block is fully
resolved; once fixed, it may lead to omissions, repetitions, or structural
errors.
We avoid this failure mode by separating parallel drafting from token
commitment.
The block-diffusion path proposes a token block in parallel, while the causal
AR path verifies the draft and determines which tokens enter the output.
The verifier accepts only the longest draft prefix that matches its
next-token predictions, preserving causal AR decoding while allowing
successful drafts to advance generation by multiple tokens.
\Cref{fig:inference} illustrates the resulting procedure.

Each round starts from the last committed token \(x_0\), followed by \(B\) mask tokens. The boundary position \(x_0\) uses causal attention, while the masked positions attend to the committed prefix and bidirectionally within the mask block. In a single forward pass, the shared decoder produces the next AR token \(a_0\) from the boundary position and the draft tokens \(d_{1:B}\) from the masked positions. Since \(a_0\) is already a causal AR prediction, it is committed directly.

To verify the draft tokens, we feed the sequence \([a_0,d_{1:B}]\) into the shared decoder under token-level causal attention. This verification pass produces the AR predictions \(a_{1:B+1}\). Following \cref{sec:prelim}, we accept the longest draft prefix \(d_{1:A}\) satisfying \(d_j=a_j\) for every \(1\leq j\leq A\), and append the next AR token \(a_{A+1}\). This final token serves as the boundary token for the next round.

\subsection{RL on the AR Path}
\label{sec:method:rl}
The joint training objective in \cref{sec:method:arloss} provides token-level
supervision for both causal AR prediction and block-diffusion drafting.
However, OCR quality is evaluated on complete outputs using sequence- and structure-level criteria that are not directly optimized by token-level likelihood training.
We therefore further optimize the jointly trained model using task-aware OCR rewards.

For multi-step masked diffusion, evaluating the sequence likelihood requires
marginalizing over unmasking trajectories, so existing RL methods approximate
it or apply policy gradients along the denoising trajectory
\citep{zhao2025d1,zhan2025agrpo}.
In \system{}, however, the causal AR verifier determines the output of
self-speculative decoding. We can therefore optimize the AR path using its
exact autoregressive likelihood and apply standard GRPO directly.
We fine-tune the full model through this path, updating the diffusion drafter
through shared parameters without a diffusion-specific objective.

\paragraph{Task-Aware OCR Rewards.}
For each sampled output, we select the reward according to its OCR task.
For plain-text recognition, we use normalized edit similarity, \(1-\mathrm{NED}\), where NED is the
normalized edit distance.
For HTML table recognition, we combine structure-only TEDS \citep{pubtabnet}
with a cell-content similarity and penalize malformed or degenerate markup.
For formula recognition, we use the edit similarity of canonicalized LaTeX,
scaled down when the output fails well-formedness checks, as a
rendering-free proxy for CDM \citep{wang2024cdm}.
All rewards lie in \([0,1]\); exact definitions and coefficients are given
in \cref{app:rl}.

\paragraph{GRPO on the Causal AR Path.}
For each input, we sample a group of outputs from the jointly trained model's causal AR path and score them with the corresponding task-aware reward. GRPO \citep{shao2024deepseekmath} computes relative advantages within each group and updates the model using causal token-level log-probabilities.
Because RL updates through the causal AR path are applied to the shared
parameters, they can also change the block-diffusion drafter and its agreement
with the verifier.
As shown in \cref{sec:exp:rl}, GRPO improves OCR accuracy while tokens per
forward remain essentially unchanged, indicating that drafter--verifier
agreement is preserved after RL.

%% file: sections/05_experiment.tex
\section{Experiments}
\label{sec:experiment}

\subsection{Setup}
\label{sec:exp:setup}

\paragraph{Model and Training.}
We initialize \system{} from the released \glmocr{} checkpoint
\citep{duan2026glmocr} and jointly fine-tune its vision encoder and language
decoder for causal AR and block-diffusion prediction.
In the original pipeline, PP-DocLayout-V3~\citep{sun2025ppdoclayout} identifies document regions, the
region-level OCR model decodes each crop, and the resulting outputs are
assembled by the merge and post-processing pipeline.
We retain the layout detector and assembly pipeline unchanged and apply the
AR-to-diffusion adaptation only to the region-level OCR model.
Unless otherwise specified, we use a block size of \(B=32\) during both
training and inference, and construct each self-speculative draft with a
single block-diffusion forward pass.
The joint objective weighs the two losses equally
(\(\lambda=1.0\)).
We train for 40{,}000 steps, processing ${\sim}26$B forward tokens in total; this count includes vision tokens and the three response streams of each example.
Further optimization and implementation details are provided in \cref{app:training}.
Unless otherwise stated, results use the final GRPO checkpoint, obtained
after 40{,}000 joint-training steps and 500 GRPO steps (\cref{sec:exp:rl}).

\paragraph{Training Data.}
The training pool contains 12.3M region-level examples assembled from predominantly public data, including DocGenome \citep{docgenome}, Docmatix \citep{docmatix},
PubTables-1M \citep{pubtables1m}, FinTabNet \citep{fintabnet},
SynthTabNet \citep{synthtabnet}, PubTabNet \citep{pubtabnet},
RVL-CDIP \citep{rvlcdip}, DocLayNet \citep{doclaynet}, and the
training split of UniMER \citep{unimernet}.
Each example is a layout region cropped from a full page at native resolution, paired with a text target.
For full-page sources, regions are detected with the same PP-DocLayout-V3 detector used at inference; table- and formula-only sources are used as whole crops.
Targets are transcriptions of each crop produced by the base \glmocr{} model, except for table crops from sources that ship cell-level annotations (about 39\% of the table stream), for which we use the original annotations converted to the evaluation markup convention.
The pool consists of three streams---page text (6.49M regions), tables
(2.90M), and formulas (2.89M)---and is predominantly English.
To realize a fixed 60/20/20 stream ratio by example count, the table and
formula streams are each subsampled once with a fixed seed to 2.16M examples,
giving a 10.8M training set that is globally shuffled.
We additionally hold out 1{,}000 pages (4{,}614 region crops) as an internal
validation set.
The pool contains no \odb{} pages (checked by document identifier, text
shingles, table cells, and formula strings); PubTabNet is used through its
training split only.

\paragraph{Benchmarks and Metrics.}
We use  \odb{} v1.6 as our primary benchmark, which contains 1,651 document pages
and evaluates document parsing under its official protocol and aggregation
\citep{omnidocbench,wang2026mineru25pro}.
We report the official Overall score together with the corresponding text,
table, formula, and reading-order metrics.
We additionally evaluate structured recognition on the PubTabNet validation
set, comprising 9,115 tables, and on UniMER-Test for mathematical expression
recognition
\citep{pubtabnet,unimernet}.
For PubTabNet, we report both full TEDS and structure-only TEDS, while UniMER
is evaluated using CDM
\citep{wang2024cdm}.
All models are evaluated using the same scoring pipeline for each benchmark.

\paragraph{Baselines.}
For in-family comparisons, we compare the original \glmocr{} model with
causal AR decoding and its built-in MTP branch
\citep{duan2026glmocr} against \system{}'s 
causal AR, direct block-diffusion, and self-speculative decoding.
Direct block-diffusion decoding serves as a diagnostic baseline that
iteratively commits confidence-selected diffusion predictions without causal
verification.
For cross-model evaluation, we report the reproduced results of
MinerU2.5 \citep{niu2025mineru25}, MinerU2.5-Pro
\citep{wang2026mineru25pro}, MinerU-Diffusion \citep{dong2026mineru},
PaddleOCR-VL-1.5 \citep{paddleocrvl15}, dots.ocr \citep{dotsocr},
DeepSeek-OCR-2 \citep{wei2026deepseekocr2}, and HunyuanOCR-1.5
\citep{li2026hunyuanocr15} using the released implementations.

\paragraph{Efficiency Measurement.}
We measure tokens per forward (TPF) over the complete decoding run as
\begin{equation}
\mathrm{TPF}
=
\frac{\text{number of committed output tokens}}
     {\text{number of forward passes}},
\end{equation}
where each block-diffusion drafting and each causal verification are counted separately as one forward pass.
TPF is measured with a single request in flight (batch size 1), so one forward
pass is one model invocation on that request, and the per-request vision and
prompt prefill forward is not counted.
We also report the number of drafted tokens accepted per round, excluding
the boundary prediction \(a_0\) and the verifier's own token, which every
round commits regardless of the draft.
Under this convention, standard AR decoding has a TPF of 1 by construction.
We additionally report decoding throughput in tokens per second (tok/s) and
document throughput in pages per second (pages/s).
Unless stated otherwise, \system{} and the \glmocr{} AR baseline are served
with SGLang \citep{zheng2024sglang} and timed through its client interface;
the \glmocr{} MTP row uses its official vLLM path, and each external system
runs on its own official inference stack (vLLM, PaddleX, or native
Transformers; see \cref{app:speed}).
The Hugging Face Transformers rows of \cref{tab:decode_modes} are an
in-process eager implementation, included as an unoptimized
reference.\footnote{Because
our training corpus is predominantly English, efficiency measurements use the
English subset of \odb{} unless otherwise stated; quality scores are computed
on the full evaluation set.}

\subsection{Main Results}
\label{sec:exp:main}

\input{tables/main_odb}
\paragraph{OmniDocBench.}
\Cref{tab:main_odb} compares \system{} with the original \glmocr{} model and
representative document parsing systems on \odb{} v1.6, reporting
recognition quality and page-processing speed side by side.
\system{} achieves an Overall score of 95.16, compared with 95.48 for the
original \glmocr{}, retaining its document parsing quality within 0.32
points. 
Relative to the base model, TEDS and CDM drop only slightly.
Apart from the original \glmocr{}, the Overall score of \system{} is
surpassed only by MinerU2.5-Pro and HunyuanOCR-1.5 among the evaluated parsers.

\paragraph{Page Processing Speed.}
The speed columns of \cref{tab:main_odb} are measured on a 100-page subset
of \odb{}, using a single H100 with one page in flight.
Each model is evaluated through its released page-processing pipeline, with
layout analysis, region recognition, and output assembly included in the
wall-clock measurement.
With self-speculative decoding, \system{} commits an average of 9.7 output
tokens per model forward and achieves 1,047 tok/s and 0.730 pages/s, the
highest processing rates in the table. Its page throughput exceeds that of
MinerU2.5-Pro at 0.399 pages/s and HunyuanOCR-1.5 with DFlash at 0.579 pages/s.
The causal AR mode of the same checkpoint reaches 794 tok/s and 0.554 pages/s,
corresponding to a \(1.32\times\) page-processing speedup from
self-speculative decoding.
\system{} also exceeds the built-in MTP mode of the original \glmocr{} at
0.472 pages/s.
Notably, the MTP mode remains slower than the original model's causal AR mode
at 0.571 pages/s despite advancing 3.7 tokens per forward, whereas the
parallelism of \system{} translates into a wall-clock page-processing gain.

\paragraph{Table and Formula Recognition.}
Beyond \odb{}, \cref{tab:main_bench} evaluates \system{} on PubTabNet and
UniMER, dedicated benchmarks for table and formula recognition, respectively.
On PubTabNet, \system{} achieves a TEDS score of 0.871 and a TEDS-struct
score of 0.916.
TEDS evaluates both table structure and cell content, whereas TEDS-struct
isolates the recovered table structure.
Compared with the original \glmocr{}, \system{} improves TEDS from 0.803 to
0.871 and TEDS-struct from 0.858 to 0.916.
On UniMER, \system{} achieves a CDM score of 0.962, closely matching the
original \glmocr{} score of 0.963 and exceeding all evaluated external
systems.

Taken together, these results show that our framework largely preserves the
recognition performance of the base model  
across the evaluated
benchmarks while enabling block-diffusion drafting and causal AR verification.

\input{tables/main_benchmarks}

\subsection{Decoding Efficiency}
\label{sec:exp:speed}

\input{tables/decode_modes}
\input{tables/diffusion_frontier}
\paragraph{Inference Backend and Measurement Scope.}
\Cref{tab:decode_modes} compares AR and self-speculative decoding under two
inference backends, Hugging Face Transformers \citep{wolf2020transformers}
and SGLang \citep{zheng2024sglang}, using the same checkpoint and region
inputs.
For each backend, we report \emph{decode-only} throughput for token generation
and \emph{end-to-end} throughput including vision encoding and prompt prefill.
With the Transformers implementation, self-speculation raises the
decode-only rate from 53 to 293 tok/s (\(5.53\times\)) and the
end-to-end rate from 51 to 161 tok/s (\(3.20\times\)).
With SGLang, self-speculation raises decode-only throughput from 777 to
3,057 tok/s (\(3.94\times\)) and end-to-end throughput from 486 to
844 tok/s (\(1.74\times\)).
The decode-only gain is larger in the Transformers implementation, whose
eager execution is bound by kernel-launch overhead: a forward costs roughly
the same whether it processes one token or a whole draft window, so the gain
approaches the tokens-per-forward ratio.
SGLang's fused kernels make each forward cheaper but scale its cost with the
number of tokens processed, and the fixed per-crop cost of vision encoding
and prompt prefill becomes a larger share of a request whose decoding is
much faster, which is why its end-to-end gain is smaller.

\input{tables/batch_throughput}
\paragraph{Direct Block-Diffusion vs.\ Self-Speculative Decoding.}
We compare direct block-diffusion and self-speculative decoding to
measure the effect of causal AR verification.
Direct block-diffusion commits confidence-selected predictions without
verification, exposing the quality--parallelism trade-off of parallel
commitment.
Self-speculative decoding instead uses the diffusion path to propose a
block and the causal AR path to determine which tokens are committed.
\Cref{fig:diffusion_frontier} compares this with MinerU-Diffusion, a
representative diffusion-native OCR model.
Across the evaluated confidence thresholds, increasing the number of
tokens committed per forward is accompanied by lower recognition accuracy
under direct diffusion decoding.
The diffusion mode of \system{} remains more accurate than
MinerU-Diffusion throughout the sweep, but exhibits the same trade-off.
At comparable parallelism of approximately 10 tokens per forward,
direct block-diffusion decoding achieves an Overall score of 92.53,
whereas self-speculative decoding achieves 95.16 at 9.7 tokens per forward.
This comparison shows that causal AR verification avoids the recognition
loss associated with direct parallel commitment while retaining comparable
parallelism.

\paragraph{Throughput Scaling with Batch Size.}
Batching can improve GPU utilization for causal AR decoding, potentially
narrowing the relative advantage of self-speculation.
\Cref{fig:batch_throughput} evaluates SGLang throughput under increasing numbers of
concurrently processed region inputs, including vision encoding,
prompt prefill, and token generation.
In this sweep, measured before RL,
self-speculative decoding leads by \(1.85\times\) at batch size one.
As the batch size increases, batching exposes more parallel work from the AR
decoder to the GPU, improving its utilization and narrowing the advantage of
self-speculation.
At batch size 64, AR decoding reaches 1,642 tok/s and self-speculative
decoding reaches 1,857 tok/s.
Although the advantage narrows at higher batch sizes, self-speculative decoding still achieves a \(1.13\times\) higher throughput.
These results show that self-speculative decoding provides its largest
throughput gains in low-concurrency settings, where sequential AR decoding
leaves more GPU capacity underutilized.

\input{tables/odb_type_accept}
\paragraph{Decoding Efficiency by Output Type.}
\Cref{tab:odb_type} shows that table regions benefit more from
self-speculative decoding than text regions. They accept 25.0 of the
32 drafted tokens per round, compared with 15.0 for text, and achieve
speedups of \(3.36\times\) and \(1.54\times\), respectively.
Table outputs are also longer, averaging 872 tokens compared with 73 for
text, which reduces the relative cost of vision encoding and prompt
prefill. Their higher draft acceptance may reflect the stronger syntactic
constraints of structured outputs.

\subsection{Effect of the AR Loss}
\label{sec:exp:arloss}

\Cref{tab:arloss_main} isolates the auxiliary AR loss.
Even without this loss, the model can perform causal verification because
it is initialized from a pretrained AR model.
Removing AR supervision lowers the Overall score from 95.02 to 93.64,
while TPF remains similar: 6.75 with the AR loss and 6.96 without it.
The AR objective therefore preserves verifier accuracy with little change
in drafting efficiency.

\input{tables/arloss_main}

\subsection{RL on the AR Path}
\label{sec:exp:rl}

\paragraph{GRPO on the AR Path.}
We evaluate whether AR-path GRPO improves OCR quality while preserving
diffusion drafting efficiency.
\Cref{tab:rl} compares direct diffusion and self-speculative decoding
before and after GRPO.

Under self-speculative decoding, GRPO improves the OmniDocBench Overall score
from 94.92 to 95.16, while TPF remains essentially unchanged at 9.61 and
9.68, respectively.
Direct diffusion also retains similar or improved accuracy and similar TPF
across the evaluated confidence thresholds, despite receiving no
diffusion-specific training objective.
These results show that AR-path GRPO improves OCR quality while preserving
drafting efficiency.

\input{tables/rl_diffusion_path}

%% file: tables/main_odb.tex
\begin{table}[t!]
\centering
\caption{\textbf{Main results on \odb{} v1.6 and page-processing speed.}
Text and Order measure normalized edit distance over character sequences
and text-block reading order, respectively; TEDS is tree-edit-distance similarity for tables, and CDM is character detection matching for formulas. Overall averages the three recognition axes.
Page speed includes each system's layout stage. TPF denotes output tokens committed per model forward.
All results are our measurements; speed protocols are in \cref{app:speed}.
$^{\dagger}$These modes use extra draft parameters; TPF counts tokens accepted per base-model verification, excluding drafting computation.}
\label{tab:main_odb}
\small
\setlength{\tabcolsep}{3.5pt}
\begin{tabular}{l l c c c c c c c c}
\toprule
 & & \multicolumn{5}{c}{\textbf{Quality}} & \multicolumn{3}{c}{\textbf{Speed}} \\
\cmidrule(lr){3-7} \cmidrule(lr){8-10}
Model & Decode & Text$^{\downarrow}$ & TEDS$^{\uparrow}$ & CDM$^{\uparrow}$ & Order$^{\downarrow}$ & Overall$^{\uparrow}$ & TPF$^{\uparrow}$ & tok/s$^{\uparrow}$ & pages/s$^{\uparrow}$ \\
\midrule
\multicolumn{10}{l}{\emph{External parsers}} \\
MinerU2.5           & AR     & 0.045 & 0.881 & 0.959 & 0.130 & 93.18 & 1.0 & 559 & 0.407 \\
MinerU2.5-Pro       & AR     & 0.037 & 0.935 & 0.969 & 0.124 & 95.57 & 1.0 & 545 & 0.399 \\
MinerU-Diffusion    & diff   & 0.073 & 0.853 & 0.916 & 0.154 & 89.87 & 5.2 & 79  & 0.059 \\
PaddleOCR-VL-1.5    & AR     & 0.042 & 0.919 & 0.969 & 0.129 & 94.86 & 1.0 & 741 & 0.389 \\
dots.ocr            & AR     & 0.048 & 0.865 & 0.917 & 0.140 & 91.14 & 1.0 & 149 & 0.116 \\
DeepSeek-OCR-2      & AR     & 0.049 & 0.859 & 0.933 & 0.144 & 91.42 & 1.0 & 24  & 0.022 \\
HunyuanOCR-1.5      & AR     & 0.036 & 0.952 & 0.949 & 0.125 & 95.52 & 1.0 & 390 & 0.313 \\
HunyuanOCR-1.5      & DFlash & 0.036 & 0.952 & 0.949 & 0.125 & 95.52 & 9.9$^{\dagger}$ & 720 & 0.579 \\
\glmocr{} (base)    & AR     & 0.040 & 0.934 & 0.970 & 0.141 & 95.48 & 1.0 & 807 & 0.571 \\
\glmocr{} (base)    & MTP    & 0.040 & 0.934 & 0.970 & 0.141 & 95.48 & 3.7$^{\dagger}$ & 667 & 0.472 \\
\midrule
\multicolumn{10}{l}{\emph{Ours}} \\
\ours{}             & AR     & 0.040 & 0.928 & 0.967 & 0.141 & 95.16 & 1.0 & 794 & 0.554 \\
\ours{}             & \textbf{self-spec} & 0.040 & 0.928 & 0.967 & 0.141 & 95.16 & \textbf{9.7} & \textbf{1{,}047} & \textbf{0.730} \\
\bottomrule
\end{tabular}
\end{table}

%% file: tables/main_benchmarks.tex
\begin{table}[t!]
\centering
\caption{\textbf{Table and formula recognition.} PubTabNet table recognition on the full validation set (9,115 tables) and UniMER formula recognition, all rows measured by us under one pipeline per benchmark, with each system's own task prompt and the same output normalisation.}
\label{tab:main_bench}
\small
\setlength{\tabcolsep}{5pt}
\begin{tabular}{l c c c}
\toprule
Model & PubTabNet TEDS$^{\uparrow}$ & TEDS-struct$^{\uparrow}$ & UniMER CDM$^{\uparrow}$ \\
\midrule
MinerU2.5             & 0.854 & 0.909 & 0.927 \\
MinerU2.5-Pro         & 0.866 & 0.913 & 0.956 \\
MinerU-Diffusion      & 0.734 & 0.858 & 0.952 \\
PaddleOCR-VL-1.5      & 0.828 & 0.900 & 0.956 \\
dots.ocr              & \textbf{0.893} & \textbf{0.935} & 0.910 \\
DeepSeek-OCR-2        & 0.849 & 0.896 & 0.863 \\
HunyuanOCR-1.5        & 0.852 & 0.903 & 0.942 \\
\glmocr{} (base, AR)  & 0.803 & 0.858 & \textbf{0.963} \\
\midrule
\ours{} & 0.871 & 0.916 & 0.962 \\
\bottomrule
\end{tabular}

\end{table}

%% file: tables/decode_modes.tex
\begin{table}[t!]
\centering
\caption{\textbf{AR vs.\ self-speculative decoding.} Decode-only and end-to-end throughput under Transformers and SGLang.}
\label{tab:decode_modes}
\small
\setlength{\tabcolsep}{5pt}
\begin{tabular}{l l c c c}
\toprule
 & & \multicolumn{2}{c}{tok/s$^{\uparrow}$} & \\
\cmidrule(lr){3-4}
Backend & Decode & decode-only & end-to-end & speedup \\
\midrule
Transformers     & AR        & 53  & 51   & --- \\
Transformers     & self-spec & 293 & 161  & 5.53$\times$ / 3.20$\times$ \\
\midrule
SGLang & AR        & 777  & 486  & --- \\
SGLang & self-spec & \textbf{3057} & \textbf{844} & 3.94$\times$ / 1.74$\times$ \\
\bottomrule
\end{tabular}
\end{table}

%% file: tables/diffusion_frontier.tex
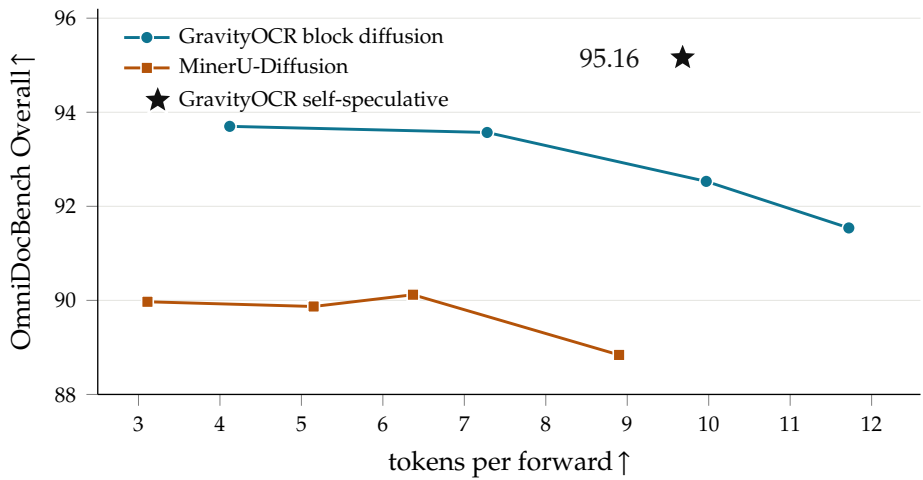
\begin{figure}[t!]
\centering
\definecolor{vizteal}{HTML}{0E7490}
\definecolor{vizamber}{HTML}{B45309}
\definecolor{vizdark}{HTML}{111111}
\definecolor{vizgrid}{HTML}{E4E4E0}
\definecolor{vizink}{HTML}{000000}
\pgfdeclareplotmark{fivestar}{%
  \node[star, star points=5, star point ratio=2.3, minimum size=12pt,
        inner sep=0pt, fill=vizdark, draw=white, line width=0.8pt] {};%
}
\begin{tikzpicture}
\begin{axis}[
  width=0.68\linewidth, height=5.1cm, scale only axis,
  xmin=2.5, xmax=12.6, ymin=88.0, ymax=96.2,
  xlabel={tokens per forward\,$\uparrow$},
  ylabel={\odb{} Overall\,$\uparrow$},
  axis x line*=bottom, axis y line*=left,
  tick align=outside,
  every tick label/.append style={font=\scriptsize, color=vizink},
  label style={font=\small, color=vizink},
  axis line style={vizink, line width=0.5pt},
  ymajorgrids, major grid style={vizgrid, line width=0.35pt},
  legend style={font=\scriptsize, draw=none, fill=none, at={(0.02,0.98)},
                anchor=north west, cells={anchor=west}},
  clip=false,
]
\addplot[vizteal, line width=1.1pt, mark=*, mark size=2.4pt,
         mark options={fill=vizteal, draw=white, line width=0.6pt}]
  coordinates {(4.12,93.70) (7.28,93.57) (9.97,92.53) (11.72,91.54)};
\addlegendentry{\ours{} block diffusion}
\addplot[vizamber, line width=1.1pt, mark=square*, mark size=2.2pt,
         mark options={fill=vizamber, draw=white, line width=0.6pt}]
  coordinates {(3.11,89.97) (5.15,89.87) (6.37,90.12) (8.90,88.84)};
\addlegendentry{MinerU-Diffusion}
\addplot[only marks, mark=fivestar] coordinates {(9.68,95.16)};
\addlegendentry{\ours{} self-speculative}
\node[font=\small, color=vizdark, anchor=east, xshift=-5pt] at (axis cs:9.44,95.16) {$95.16$};
\end{axis}
\end{tikzpicture}
\caption{\textbf{Quality and parallelism of diffusion decoding.} Curves sweep each system's confidence threshold \(\tau_c\). Direct block-diffusion TPF counts forwards that commit tokens, excluding cache writes; self-speculative TPF counts both drafting and verification (\cref{app:speed}).}
\label{fig:diffusion_frontier}
\end{figure}

%% file: tables/batch_throughput.tex
\begin{figure}[t!]
\centering
\definecolor{vizblue}{HTML}{2A78D6}
\definecolor{vizorange}{HTML}{EB6834}
\definecolor{vizgrid}{HTML}{E4E4E0}
\definecolor{vizink}{HTML}{000000}
\begin{tikzpicture}
\begin{axis}[
  width=0.66\linewidth, height=4.6cm, scale only axis,
  xmode=log, log basis x=2,
  xmin=0.85, xmax=78, ymin=350, ymax=2000,
  xtick={1,2,4,8,16,32,64},
  xticklabels={1,2,4,8,16,32,64},
  ytick={500,1000,1500,2000},
  xlabel={batch size},
  ylabel={throughput (tok/s)\,$\uparrow$},
  axis x line*=bottom, axis y line*=left,
  tick align=outside,
  every tick label/.append style={font=\scriptsize, color=vizink},
  label style={font=\small, color=vizink},
  axis line style={vizink, line width=0.5pt},
  ymajorgrids, major grid style={vizgrid, line width=0.35pt},
  legend style={font=\scriptsize, draw=none, fill=none, at={(0.03,0.97)},
                anchor=north west, cells={anchor=west}},
  clip=false,
]
\addplot[vizorange, line width=1.1pt, mark=*, mark size=2.2pt,
         mark options={fill=vizorange, draw=white, line width=0.6pt}]
  coordinates {(1,948.4) (2,1155.7) (4,1394.7) (8,1600.1) (16,1759.7) (32,1831.3) (64,1857.2)};
\addlegendentry{self-spec (ours)}
\addplot[vizblue, line width=1.1pt, mark=*, mark size=2.2pt,
         mark options={fill=vizblue, draw=white, line width=0.6pt}]
  coordinates {(1,513.5) (2,761.8) (4,1012.4) (8,1249.9) (16,1412.9) (32,1484.3) (64,1641.9)};
\addlegendentry{AR}
\draw[<->, vizink, line width=0.5pt]
  (axis cs:1,513.5) -- (axis cs:1,948.4);
\node[font=\scriptsize\bfseries, color=vizink, anchor=west, xshift=3pt]
  at (axis cs:1,731) {$1.85\times$};
\draw[<->, vizink, line width=0.5pt]
  (axis cs:64,1641.9) -- (axis cs:64,1857.2);
\node[font=\scriptsize\bfseries, color=vizink, anchor=west, xshift=3pt]
  at (axis cs:64,1750) {$1.13\times$};
\end{axis}
\end{tikzpicture}
\caption{\textbf{Throughput vs.\ batch size.} The speedup narrows with batch size: speculation leads by $1.85\times$ at batch 1 and by $1.13\times$ at batch 64. Protocol in \cref{app:speed}.}
\label{fig:batch_throughput}
\end{figure}
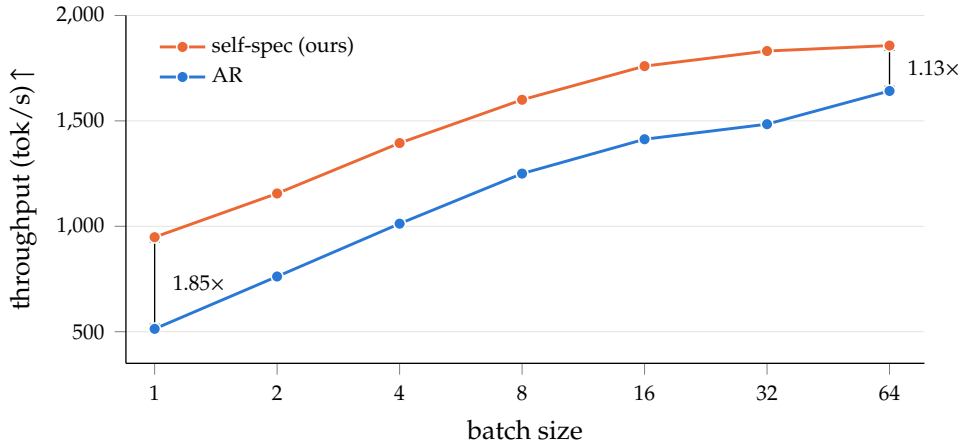

%% file: tables/odb_type_accept.tex
\begin{table}[t!]
\centering
\caption{\textbf{Parallelism and speed by content type.} Accepted tokens is the number of draft tokens accepted per round. TPF divides committed output tokens by all draft and verification forwards pooled within each content type.}
\label{tab:odb_type}
\small
\setlength{\tabcolsep}{5pt}
\begin{tabular}{l r r c c c c c}
\toprule
 & & mean output & accepted & & \multicolumn{2}{c}{tok/s$^{\uparrow}$} & \\
\cmidrule(lr){6-7}
Type & crops & tokens & tokens$^{\uparrow}$ & TPF$^{\uparrow}$ & AR & self-spec & speedup \\
\midrule
Text    & 7{,}019 & 73  & 15.0 & 8.5 & 419 & 647  & 1.54$\times$ \\
Formula & 1{,}660 & 95  & 18.9 & 10.5 & 520 & 947  & 1.82$\times$ \\
Table   & 243     & 872 & \textbf{25.0} & \textbf{13.5} & 732 & \textbf{2{,}462} & \textbf{3.36$\times$} \\
\midrule
All     & 8{,}922 & 99 & 17.4 & 9.7 & 486 & 844  & 1.74$\times$ \\
\bottomrule
\end{tabular}
\end{table}

%% file: tables/arloss_main.tex
\begin{table}[t!]
\centering
\caption{\textbf{Effect of the auxiliary AR loss on self-speculative decoding.} Both models are 10k-step checkpoints that share the training setup except the AR loss weight.}
\label{tab:arloss_main}
\small
\setlength{\tabcolsep}{6pt}
\begin{tabular}{l c c}
\toprule
 & \odb{} Overall$^{\uparrow}$ & TPF$^{\uparrow}$ \\
\midrule
no AR loss & 93.64 & 6.96 \\
AR loss    & \textbf{95.02} & 6.75 \\
\bottomrule
\end{tabular}
\end{table}

%% file: tables/rl_diffusion_path.tex
\begin{table}[t!]
\centering
\caption{\textbf{GRPO on the AR path.}
\odb{} v1.6 Overall and TPF for direct block-diffusion and self-speculative
decoding before and after GRPO.}
\label{tab:rl}
\small
\setlength{\tabcolsep}{4pt}
\begin{tabular}{l l c c c c c}
\toprule
 & & \multicolumn{3}{c}{Overall$^{\uparrow}$} & \multicolumn{2}{c}{TPF$^{\uparrow}$} \\
\cmidrule(lr){3-5}\cmidrule(lr){6-7}
Decode & $\tau_c$ & before & after & $\Delta$ & before & after \\
\midrule
Block diffusion & $0.7$  & 86.17 & 86.06 & $-0.11$ & 14.85 & 15.43 \\
Block diffusion & $0.9$  & 91.55 & 91.54 & $-0.01$ & 11.89 & 11.72 \\
Block diffusion & $0.95$ & 89.62 & \textbf{92.53} & $\mathbf{+2.91}$ & 10.20 & 9.97 \\
Block diffusion & $0.99$ & 93.23 & 93.57 & $+0.34$ & 7.18 & 7.28 \\
\midrule
Self-spec & ---    & 94.92 & \textbf{95.16} & $\mathbf{+0.24}$ & 9.61 & 9.68 \\
\bottomrule
\end{tabular}
\end{table}

%% file: sections/06_related_work.tex
\section{Related Work}
\label{sec:related_work}

\paragraph{Document Parsing VLMs.}
Generative document parsing models translate document images into serialized
text and structured markup.
Early systems such as Nougat focused on image-to-markup transcription for
scientific documents, while olmOCR extended generative parsing to large-scale
document conversion
\citep{nougat,olmocr}.
More recent OCR-specialized VLMs, including \glmocr{}, MinerU2.5-Pro,
OvisOCR2, PaddleOCR-VL, dots.ocr, and DeepSeek-OCR, improve document parsing
through compact architectures, specialized training data, and document-aware
visual processing
\citep{duan2026glmocr,wang2026mineru25pro,lu2026ovisocr2,paddleocrvl,
dotsocr,deepseekocr}.
For example, DeepSeek-OCR compresses document images into compact optical
contexts, PaddleOCR-VL adopts coarse-to-fine visual processing, and
\glmocr{} combines layout analysis with region-level content generation.
These studies primarily improve visual representation, training data, or the
document-processing pipeline, while retaining causal AR generation for the
output sequence.

A smaller body of work directly targets OCR decoding efficiency.
\glmocr{} incorporates an MTP branch that predicts multiple future tokens
from its AR backbone
\citep{duan2026glmocr}.
Because this branch is included in our base OCR system, we use it as the
direct in-model acceleration baseline.
HunyuanOCR-1.5 instead adapts DFlash to OCR by training a separate
block-diffusion drafter conditioned on features from the target AR model
\citep{li2026hunyuanocr15,chen2026dflash}.
HSD accelerates document parsing at a different granularity: it constructs
region-level drafts, verifies regions in parallel, and performs an additional
page-level verification stage to preserve global document coherence
\citep{liao2026hsd}.
These approaches establish multi-token drafting and verification as practical
directions for accelerating generative OCR, but differ in whether drafting is
performed by auxiliary prediction heads, an external draft model, or a
hierarchical document pipeline.

\paragraph{Diffusion Language Models.}
Discrete diffusion modeling for text originated from denoising objectives
over corrupted token sequences
\citep{austin2021d3pm,sahoo2024mdlm}, and was subsequently scaled to large
diffusion language models such as LLaDA and Dream
\citep{nie2025llada,ye2025dream}.
Later work has focused on making diffusion generation compatible with
efficient language-model serving.
Block Diffusion introduces a block-wise formulation that supports reuse of
completed context
\citep{arriola2025block}.
Fast-dLLM develops caching-aware inference techniques for pretrained
diffusion language models
\citep{wu2025fastdllm}, while Fast-dLLM v2 proposes a data-efficient
AR-to-block-diffusion adaptation scheme using complementary clean and
corrupted training streams
\citep{wu2025fastdllmv2}.
WeDLM follows a different approach, using causal attention and topological
token reordering to enable parallel prediction while retaining compatibility
with standard prefix KV caching
\citep{liu2025wedlm}.

Diffusion decoding has recently been introduced to document OCR.
DODO develops a block-discrete diffusion OCR model designed to mitigate the
structural instability of global masked diffusion
\citep{man2026dodo}.
MinerU-Diffusion formulates document OCR as inverse rendering and combines a
block-wise diffusion decoder with an uncertainty-driven training curriculum
\citep{dong2026mineru}.
These models demonstrate that document outputs can be recovered through
parallel diffusion decoding and provide favorable accuracy--efficiency
trade-offs.
Nevertheless, direct diffusion decoding has not consistently matched the
recognition accuracy of strong AR OCR systems across all reported benchmarks
and operating points, motivating methods that retain parallel proposals
without directly committing every diffusion prediction.

\paragraph{AR--Diffusion Hybrid Models.}
Rather than replacing pretrained AR models with diffusion-only models, recent
work adapts AR checkpoints to support both causal and diffusion-style
prediction.
SDAR studies data-efficient conversion from pretrained AR models to
block-wise diffusion models
\citep{cheng2025sdar}.
Fast-dVLM extends direct AR-to-diffusion conversion to vision--language
models, jointly training causal and denoising objectives so that the adapted
model retains both generation modes
\citep{wu2026fastdvlm}.
Nemotron-Labs-Diffusion scales this formulation to a family of language and
vision--language models that can operate in AR, diffusion, or
self-speculative modes
\citep{fu2026nemotron}.
TiDAR more tightly integrates the two modes through a structured attention
pattern that combines diffusion-based drafting with AR generation
\citep{liu2025tidar}.

A related line of work uses diffusion models specifically as drafters for an
AR target.
Speculative diffusion decoding and DFlash train dedicated diffusion drafters
for parallel block proposal
\citep{christopher2024speculative,chen2026dflash}.
DiffuSpec instead repurposes a pretrained diffusion language model as a
training-free drafter, while SpecDiff-2 improves agreement between the
diffusion drafter and AR verifier
\citep{li2025diffuspec,sandler2025specdiff2}.
A further line studies how to align a block drafter with left-to-right
verification, through position-aware loss weighting, accept-until-fail
supervision, or targeted repair of uncertain positions
\citep{wu2026dpace,whalen2026teaching,yang2026specauf,liu2026cure}.
Across these approaches, speculative verification allows diffusion prediction
errors to reduce accepted draft length rather than directly alter the output
of the AR target.
Our method follows the shared-model branch of this literature: we adapt the
region-level generation model of \glmocr{} to support block-diffusion
drafting and causal AR verification within the same parameter set, without
introducing a separate draft network.

\paragraph{Reinforcement Learning for Diffusion Language Models.}
Policy-gradient RL over a diffusion language model requires a sequence
likelihood that depends on the denoising trajectory, so existing methods
either optimize an approximation of it or derive step-level gradients that
remain unbiased \citep{zhao2025d1,zhan2025agrpo}.
Self-speculative decoding has separately been used to accelerate the rollout
phase of RL itself \citep{kim2026efficientrollout}.
Our setting avoids the estimation problem rather than solving it: because the
adapted model retains an exact causal factorization, GRPO
\citep{shao2024deepseekmath} can be applied through the AR path, and the
shared parameters it updates are the ones the diffusion drafter uses.

%% file: sections/07_conclusion.tex
\section{Conclusion}
\label{sec:conclusion}

We presented \system{}, a parameter-shared framework that combines
block-diffusion drafting with causal AR verification to accelerate document OCR.
Joint AR--diffusion training enables a single model to draft token blocks
in parallel and verify them before commitment, preserving its AR output
while allowing multiple tokens to be committed per round.
We further apply GRPO through the causal AR path using sequence- and
structure-level OCR rewards, avoiding diffusion-trajectory likelihood
estimation while updating the shared drafter and verifier parameters.
On \odb{} v1.6, this improves the Overall score from 94.92 to 95.16,
close to the original \glmocr{} score of 95.48, while preserving
diffusion drafting efficiency.
In SGLang, \system{} achieves a \(3.94\times\) decode-only speedup
on region crops and a \(1.32\times\) end-to-end page-processing
speedup over AR decoding.
These results show that parameter-shared drafting and verification
accelerate document OCR while largely retaining the base model's
recognition capabilities, without a separate drafter network.

%% file: sections/appendix.tex
\section{Training Details}
\label{app:training}

\Cref{tab:hparams} lists the training configuration.
All parameters are trained, including the vision encoder, which uses
\(0.1\times\) the decoder learning rate; the only architectural addition is
the \(\mathtt{[M]}\) embedding, a new vocabulary row initialized by
mean-resizing, i.e.\ drawn from a normal distribution fitted to the existing
embedding statistics.
The joint loss is normalized so that its coefficients sum to one,
\(0.5\,\mathcal{L}_{\mathrm{diff}}+0.5\,\mathcal{L}_{\mathrm{AR}}\)
for \(\lambda=1\), leaving learning-rate semantics unchanged.

\begin{table}[t!]
\centering
\caption{\textbf{Training configuration.}}
\label{tab:hparams}
\small
\setlength{\tabcolsep}{6pt}
\begin{tabular}{l l}
\toprule
Hardware & 16$\times$H100 (2 nodes) \\
Steps & 40{,}000 (warmup 500, linear decay to 0) \\
Optimizer & AdamW (\(\beta_1{=}0.9\), \(\beta_2{=}0.999\)), no weight decay \\
Peak LR & \(2\times10^{-5}\) decoder, \(2\times10^{-6}\) vision encoder \\
Precision / parallelism & bf16, ZeRO stage 2, gradient clipping 1.0 \\
Batch & 1 packed row/GPU $\times$ 16 GPUs $\times$ accumulation 4 \\
Packing budget & 10{,}240 forward tokens/row (response cap 3{,}072) \\
Block size / mask schedule & \(B{=}32\), \(t\sim\mathcal{U}(0,1)\) \\
Loss weights & \(0.5\,\mathcal{L}_{\mathrm{diff}}+0.5\,\mathcal{L}_{\mathrm{AR}}\) \\
\bottomrule
\end{tabular}
\end{table}

\paragraph{Shared-prefix packing.}
Because the three response streams of \cref{fig:training} share one image
and prompt, a training row costs
\(\mathrm{prompt}+\mathrm{vision}+3\cdot\mathrm{response}\) forward tokens,
and the packer fills each row against this cost rather than the raw sequence
length---the same consideration behind Fast-dVLM's vision-efficient
concatenation \citep{wu2026fastdvlm} and Nemotron's asymmetric dual stream
\citep{fu2026nemotron}.
Rows are filled first-fit with carry-over: a sample that does not fit starts
the next batch, and typical rows pack 21--30 document samples.

\paragraph{EOS block fill.}
Each response is padded with exactly \(B\) EOS tokens during training, so
the model learns to fill the tail block with an EOS run instead of leaking
length information through block alignment.
The AR loss counts only the first EOS of each run, while the diffusion
streams supervise all of them; without this restriction the AR path
over-predicts EOS and Overall drops by more than four points.

\section{Decoding Details and Equality Testing}
\label{app:decoding}

\subsection{Direct Block-Diffusion Decoding}
\label{app:decoding:diffusion}

The completed prefix is represented by causal KV states, and generation
proceeds block by block.
Each block is initialized with \(B\) \(\mathtt{[M]}\) tokens; under the
shifted convention, the prediction for the block's first position comes from
the final position of the preceding block.

At each denoising step, one forward pass predicts all remaining masked
positions in parallel, reading the distribution for position \(p\) from the
logit at position \(p-1\).
Every prediction whose token probability exceeds the confidence threshold
\(\tau_c\) is revealed; if none exceeds it, the single most confident
position is revealed instead.
Revealed tokens remain fixed in later steps.

Once the block is complete, one additional forward pass writes its KV states
into the cache.
This pass extends the cache only and does not verify or replace any token.
Generation stops at the block containing the EOS token, and the output is
truncated at the first EOS.

\subsection{Self-Speculative Decoding and KV-Cache Update}
\label{app:decoding:selfspec}

One self-speculative round of the serving implementation proceeds as
follows; \cref{fig:inference} illustrates the same round.

\begin{enumerate}
\item \textbf{State.} The KV cache holds causal states for every committed
token except the boundary token \(x_0\), the last committed token, which was
produced as the verifier's prediction in the preceding round and has not yet
been fed to the model as input (the initial boundary is the first token
generated during prefill).
\item \textbf{Draft forward.} A window is formed from \(x_0\) followed by
\(B\) \(\mathtt{[M]}\) tokens.
The boundary position attends causally to the cache; the masked positions
attend causally to the cache and bidirectionally within the window.
One forward pass yields, under the shifted alignment, \(a_0\) from the
boundary logit, which is identical to the causal AR prediction, and the
draft \(d_{1:B}\) from the mask logits.
No confidence threshold is applied: the one-shot draft proposes all
positions unconditionally.
\item \textbf{Verify forward.} The sequence \(a_0, d_{1:B}\) is processed
under token-level causal attention in a single forward pass, attending to
the cache and to \(x_0\), whose causal KV state was written by the draft
forward, and producing the AR predictions \(a_{1:B+1}\).
The verification pass reuses the draft's KV slots for the mask positions and
overwrites their bidirectional states with causal ones, so the cache that
later rounds read is always causal.
\item \textbf{Accept.} \(A\leq B\) is the length of the longest prefix
with \(d_j=a_j\).
The round commits \(a_0\), \(d_{1:A}\), and \(a_{A+1}\); the committed
tokens are the verifier's own predictions at every position.
\item \textbf{Cache update.} The causal KV states of \(x_0\), \(a_0\) and
\(d_{1:A}\) are retained; the states of the rejected suffix are freed.
No bidirectional draft state survives the round.
\item \textbf{Next round.} \(a_{A+1}\) becomes the next boundary token and
obtains its KV state in the next draft forward.
If \(A=B\), the draft was fully accepted and \(a_{A+1}\) is the bonus
prediction following the entire block.
If any committed token completes a stop condition, the committed run is
truncated at that token and the remainder of the round is discarded.
\end{enumerate}

A round therefore commits between two tokens (\(a_0\) and \(a_1\)) and
\(B{+}2\) tokens (\(a_0\), the full draft, and the bonus token).

\subsection{AR-Equivalence and Equality Testing}
\label{app:decoding:equality}

\paragraph{Argument.}
Under top-\(1\) decoding, every committed token is the verifier's argmax
prediction given the committed prefix: \(a_0\) is computed under strictly
causal attention at the boundary, each accepted draft token satisfies
\(d_j=a_j\) by the acceptance rule, and \(a_{A+1}\) is the verifier's own
prediction.
By induction over rounds, the committed sequence equals the sequence that
standalone greedy AR decoding produces from the same checkpoint, in exact
arithmetic.

\paragraph{Empirical check.}
We compare the decoded output strings of the two decoders under identical
checkpoint, prompts, region crops, and greedy decoding: the final GRPO
checkpoint, both decoders served with SGLang, block size \(B{=}32\),
repetition penalty \(1.0\), and no output cap.
On the full English \odb{} crop set (8{,}922 crops), the two outputs are
identical on \(96.6\%\) of crops.
The remaining differences are consistent with near-tie argmax flips under
bf16 kernels: the AR and verification forwards use different attention
kernels, and at the first divergent position the median top-two logit
margin is \(0.14\) nats.
Re-evaluating the 305 first divergences in fp32, the fp32 argmax agrees
with the AR token in 228 cases and with the self-speculative token in 77,
and 65 of these positions are ties with a margin below \(0.05\) nats.

\section{Speed-Measurement Protocol and Profiles}
\label{app:speed}

\paragraph{What each speed table measures.}
All speed tables and Figure~\ref{fig:batch_throughput} use one H100, full
resolution, no output cap (natural EOS), and are run solo (no co-tenant job
on the node).

\emph{Speed columns of Table~\ref{tab:main_odb} --- page-level, single stream.} The unit is one
\odb{} page and pages are submitted \textbf{one at a time} ($C{=}1$). The wall
clock covers the \textbf{entire pipeline including layout analysis}
(112--146\,ms/page for the GLM-family systems, measured live inside the
wall) plus region cropping, recognition and markdown assembly. Systems differ
in what happens \emph{inside} one page, and we keep each system's own design:
region-pipeline models (ours, GLM-OCR) detect ${\sim}18$ regions and fire
those recognition requests concurrently, MinerU runs its own two-step pipeline
with a much heavier layout stage (${\sim}1.3$\,s/page), while full-page models
(HunyuanOCR, dots.ocr, DeepSeek-OCR-2) answer the page in a single request and
have no layout stage. Each competitor runs on its own official inference stack: HunyuanOCR-1.5
on vLLM \citep{kwon2023vllm} with the vendor recipe, MinerU2.5 and
MinerU2.5-Pro through their native pipeline over a vLLM engine, dots.ocr on
a vLLM server, PaddleOCR-VL-1.5 on the PaddleX generative-AI server with a
vLLM backend, DeepSeek-OCR-2 and MinerU-Diffusion through their native
Transformers and diffusion pipelines, and the \glmocr{} MTP mode on vLLM;
the \glmocr{} AR baseline and our model are served with SGLang.

\emph{Concurrency regimes.} Same client, pages and layout-included boundary
as Table~\ref{tab:main_odb}, measured before RL; the
\textbf{only} change is how the ${\sim}18$ region requests of a page are
issued: strictly sequentially ($C{=}1$: self-spec $1.54\times$ over AR,
$0.420$ vs $0.273$ pages/s) or all at once ($C{=}18$: $1.34\times$, $0.781$
vs $0.581$). This isolates how much of the speculative advantage comes from
an otherwise idle GPU.

\emph{Table~\ref{tab:decode_modes} --- crop-level, batch 1, no layout.} The
unit is a single ground-truth region crop from the English \odb{} set (8{,}922
crops), fed directly to the model, so \textbf{layout is not involved at all}
and there is no page assembly. Here \emph{end-to-end} is the full request wall
(vision encoding $+$ prefill $+$ decode); \emph{decode-only} counts only
the time spent generating output tokens, excluding the per-crop vision
encoding and prompt prefill.

\emph{Figure~\ref{fig:batch_throughput} --- crop-level throughput with
varying batch size.} Throughput is measured before RL on layout-detected
crops from 755 English \odb{} pages.
Layout is run once beforehand and excluded from timing.
We measure the time to process all crops, including vision encoding,
prompt prefill, and decoding, with a full client queue
($C=\max(64,\,4\times\mathrm{bs})$) and the server's running-request cap
set to the plotted batch size.
Table~\ref{tab:decode_modes} instead uses crops defined by the dataset's
annotated region boundaries, submitted one at a time, and the final GRPO
checkpoint.

\paragraph{Forward-count accounting.}
For \glmocr{} MTP and DFlash, TPF counts base-model verification forwards
and excludes auxiliary drafting computation.
For \system{} self-speculation, it counts both full-model draft and
verification forwards, excluding prefill.
For direct block-diffusion decoding, it counts only forwards that commit at
least one token, excluding prefill and the cache-write forward performed
after each completed 32-token block.
This direct-diffusion convention matches the step-count accounting used for
MinerU-Diffusion.

\section{Additional Results}
\label{app:results}

\paragraph{Speedup by Output Length.}
\Cref{tab:length_domain} breaks the crop-level comparison of
\cref{tab:decode_modes} down by output length.
Both acceptance and speedup rise monotonically with length, from
\(1.43\times\) below 128 output tokens to \(3.68\times\) above 1{,}024.
Two effects compound.
Longer outputs spend more of their wall-clock time in decoding, which is the
only stage speculation accelerates, so the fixed per-crop vision and prefill
cost stops diluting the gain.
At the same time, acceptance itself grows with length, from 14.8 of 32
drafted tokens per round below 128 output tokens to 23.9 above 1{,}024---long regions are
dominated by structured, syntax-constrained spans that the one-shot draft
predicts well---so each verification also commits more tokens.
Four fifths of \odb{} crops fall in the shortest bin, which is why the
crop-averaged speedup of \(1.74\times\) sits well below the longest bin;
the content-type breakdown in \cref{tab:odb_type} is largely this length
effect in disguise, with table regions both the longest and the
fastest-accelerating outputs.

\input{tables/length_domain_speed}

\paragraph{Draft Schedule.}
\Cref{fig:draftsweep} sweeps the number of denoising steps used to build
the draft, controlled by the draft-side confidence threshold \(\tau_d\), measured
on 400 English \odb{} crops.
Refining the draft over more steps raises the accepted length per round
(\cref{fig:draftsweep:accept}) but costs additional draft forwards, and the
extra forwards cost more than the tokens they buy: tokens per forward falls
monotonically as the draft is refined (\cref{fig:draftsweep:tpf}).
One-shot drafting is therefore the operating point we deploy.

\input{figures/draft_threshold_sweep}

\paragraph{Direct-Diffusion Quality.}
\Cref{tab:rl} also reports the diagnostic direct-diffusion mode of
\cref{app:decoding:diffusion} on the official protocol at four confidence
thresholds.
After GRPO, quality rises monotonically with \(\tau_c\) while parallelism falls; even the best
direct-diffusion operating point trails self-speculative decoding on both
axes simultaneously.

\paragraph{Mask Schedule.}
\Cref{tab:fullmask} compares the uniform partial-masking objective against
an all-mask variant matched in every other training setting, evaluated at
the same step counts on the self-speculative path.
The all-mask model is behind on quality at five of six checkpoints and commits fewer tokens per forward, indicating that the partial-masking objective, not merely exposure to masked positions, is what makes the draft strong.

\input{tables/fullmask_vs_diffusion}

\paragraph{The \(\tau_c{=}0.95\) Gap After GRPO.}
In \cref{tab:rl}, direct diffusion gains \(+2.91\) Overall at
\(\tau_c{=}0.95\) while every other threshold moves by less than \(0.4\).
The pre-GRPO model degrades three times as many formulas at exactly that
setting---116 versus 37 expressions scoring zero CDM out of 2{,}352---which
is also why its accuracy is non-monotone in \(\tau_c\) while the post-GRPO
curve is not.

\section{RL Details}
\label{app:rl}

\paragraph{Optimization.}
We apply GRPO with full fine-tuning of the entire model (no LoRA), starting
from the jointly trained 40k checkpoint.
Each step draws 24 prompts with 28 rollouts per prompt (672 completions);
advantages are standardized within each group, the loss is token-level, and
clipping is asymmetric (\(\varepsilon_{\mathrm{low}}{=}0.2\),
\(\varepsilon_{\mathrm{high}}{=}0.28\)).
A KL penalty with \(\beta{=}10^{-3}\) anchors the policy to a frozen copy of
the initial checkpoint.
Training uses a learning rate of \(3\times10^{-6}\) with cosine decay,
8-bit AdamW, bf16, and gradient clipping 1.0; checkpoints are saved every
250 steps, and we report step 500 (about 1.6 epochs over the prompt pool).
Truncated completions are masked out of the loss.
No diffusion-specific objective is applied during GRPO; the drafter changes
only through the shared parameters updated by the AR-path loss.
On the 8{,}922 English \odb{} crops, tokens per forward are 9.61 before RL
and 9.68 at step 500.

\paragraph{Rollouts.}
Rollouts are sampled from the causal AR policy at temperature 1.0 without
nucleus truncation, with a maximum completion length of 8{,}448 tokens,
using a colocated vLLM engine \citep{kwon2023vllm}.
Policy log-probabilities are recomputed by the training framework through
the same causal factorization, with token-level importance-sampling
correction between the rollout and training kernels.

\paragraph{Rewards.}
All rewards lie in \([0,1]\); CDM is not used as a reward.
Both the prediction and the reference are first normalized with the same
markup normalization the official \odb{} scorer applies, so the reward
cannot be improved by markup conventions the evaluation ignores.
Plain text is scored by the normalized Levenshtein similarity of the two
normalized strings.
For tables, let \(\mathrm{TEDS}_s\) be the structure-only TEDS and \(s_{\mathrm{cell}}\) the
normalized Levenshtein similarity of the concatenated cell strings in
reading order; the reward is
\begin{align*}
r_{\mathrm{table}} &= \mathrm{clip}_{[0,1]}\big((0.45\,\mathrm{TEDS}_s + 0.55\,s_{\mathrm{cell}})\,(1-p_{\mathrm{row}})\,(1-p_{\mathrm{loop}})\big),\\
p_{\mathrm{row}} &= 0.35\,\min(1,\mathrm{over}/d) + 0.35\,\min(1,\mathrm{under}/d),\qquad d=\max(4,\,n_{\mathrm{gt}}),\\
p_{\mathrm{loop}} &= \begin{cases} 0 & f = 0,\\ \min(1,\;0.30+0.70\,f) & f > 0,\end{cases}
\end{align*}
where \(\mathrm{over}\) and \(\mathrm{under}\) are the row-count excess and
shortfall relative to the \(n_{\mathrm{gt}}\) reference rows, and \(f\) is
the fraction of predicted rows that are excess repeats of a row key beyond
\(\max(3,\,n_{\mathrm{gt}}(\mathrm{key})+1)\), zero if no row repeats that
often.
The result is then raised toward one by an exact-grid bonus,
\(r\leftarrow r+b\,(1-r)\), where \(b\) is \(0.05\) if the predicted row
count matches the reference plus \(0.05\) if the predicted column count
matches.
An unclosed \(\texttt{<table>}\) scores zero (the evaluator drops such
tables), and a table that cannot be parsed scores \(0.15\) times the plain
edit similarity.
For formulas, the reward is
\(r_{\mathrm{formula}}=\mathrm{sim}\big(\mathrm{canon}(\hat{y}),\mathrm{canon}(y)\big)\cdot 0.3^{\,v}\),
where \(\mathrm{sim}\) is normalized Levenshtein similarity and
\(\mathrm{canon}\) strips math delimiters, maps \texttt{\textbackslash dfrac}
and \texttt{\textbackslash tfrac} to \texttt{\textbackslash frac}, and removes
\texttt{\textbackslash left}/\texttt{\textbackslash right}, spacing commands,
whitespace, and braces; \(v\) counts the well-formedness checks
(\texttt{\textbackslash left}/\texttt{\textbackslash right} balance, brace
balance, \texttt{\textbackslash begin}/\texttt{\textbackslash end}
environments, and unescaped \texttt{\$} parity) failed by the prediction but
not by the reference.
Every reward is finally multiplied by a degeneration guard
\(g=\min\big(\rho_8,\;\max(0,\,1-(L_{\mathrm{run}}-20)/200)\big)\), where
\(\rho_8\) is the distinct 8-gram ratio, computed only for outputs of at
least 16 whitespace-separated tokens and taken as one otherwise, and
\(L_{\mathrm{run}}\) is the longest single-character run.

\paragraph{Prompt pool.}
The pool contains 7{,}723 region crops from public document datasets of the
same kind as the training pool; it contains no benchmark images, and the
PubTabNet portion uses its training split only.
Its composition is 92\% tables, 6\% text, and 2\% formulas.
References are the source datasets' original annotations for the table and
formula crops (94\% of the pool) and teacher transcriptions for the text
crops.
Candidate prompts were screened by sampling eight rollouts each and
discarding prompts whose reward variance is zero (their group advantage
vanishes under GRPO), and the remainder was mined toward hard and long
tables; the median reference length is about 1{,}900 characters.

\section{Qualitative Examples}
\label{app:qualitative}
\begin{figure}[p]
  \centering
  \includegraphics[width=0.98\linewidth,height=0.9\textheight,keepaspectratio]{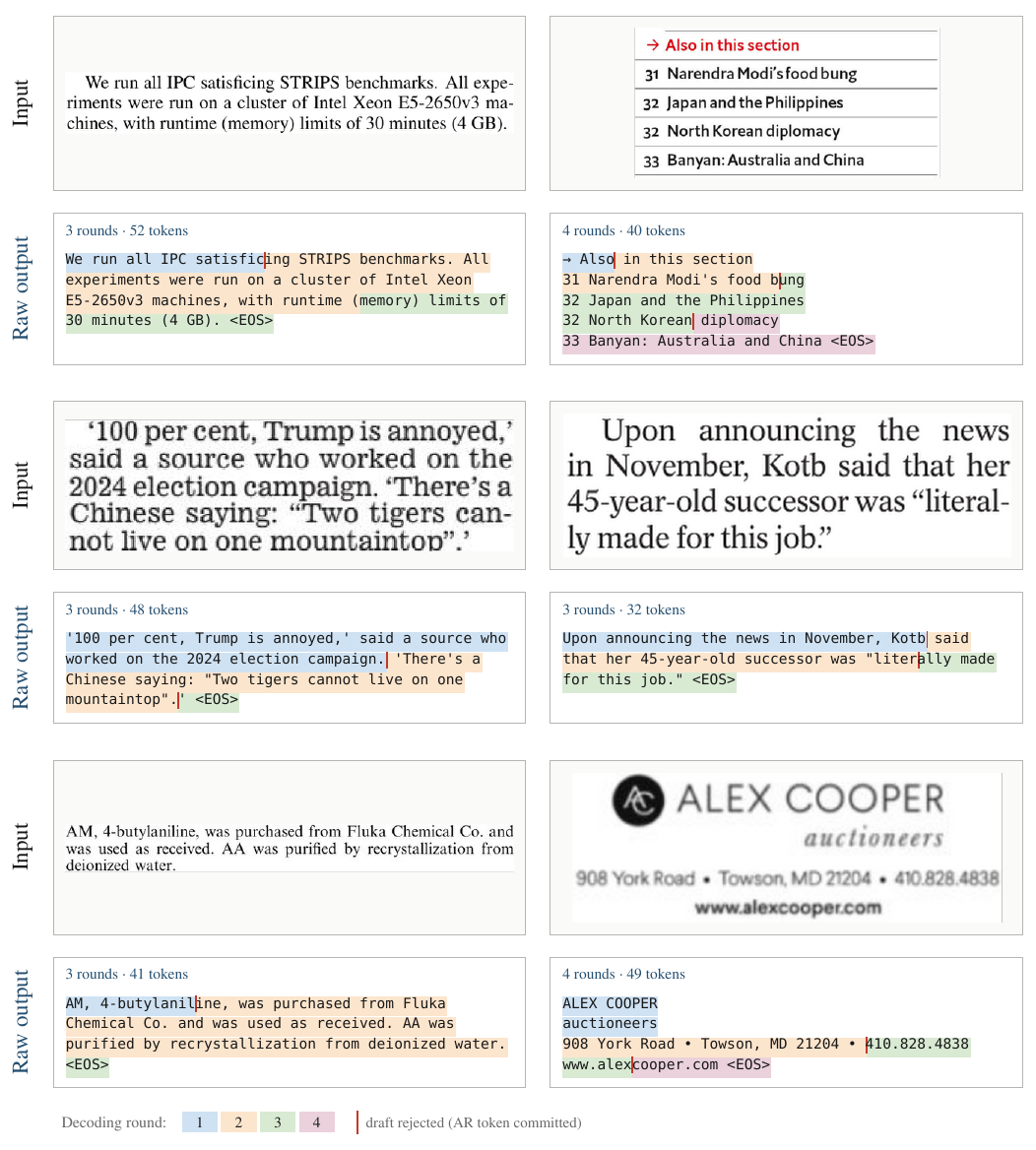}
  \caption{\textbf{Qualitative results on text regions.} Self-speculative decoding on OmniDocBench.
  Each block shows the input crop (top) and the raw output (bottom); each color denotes a different
  decoding round, a red tick marks a rejected draft token, and \texttt{<EOS>} is the end-of-sequence
  token.}
  \label{fig:qual-text}
\end{figure}

\begin{figure}[p]
  \centering
  \includegraphics[width=0.98\linewidth,height=0.9\textheight,keepaspectratio]{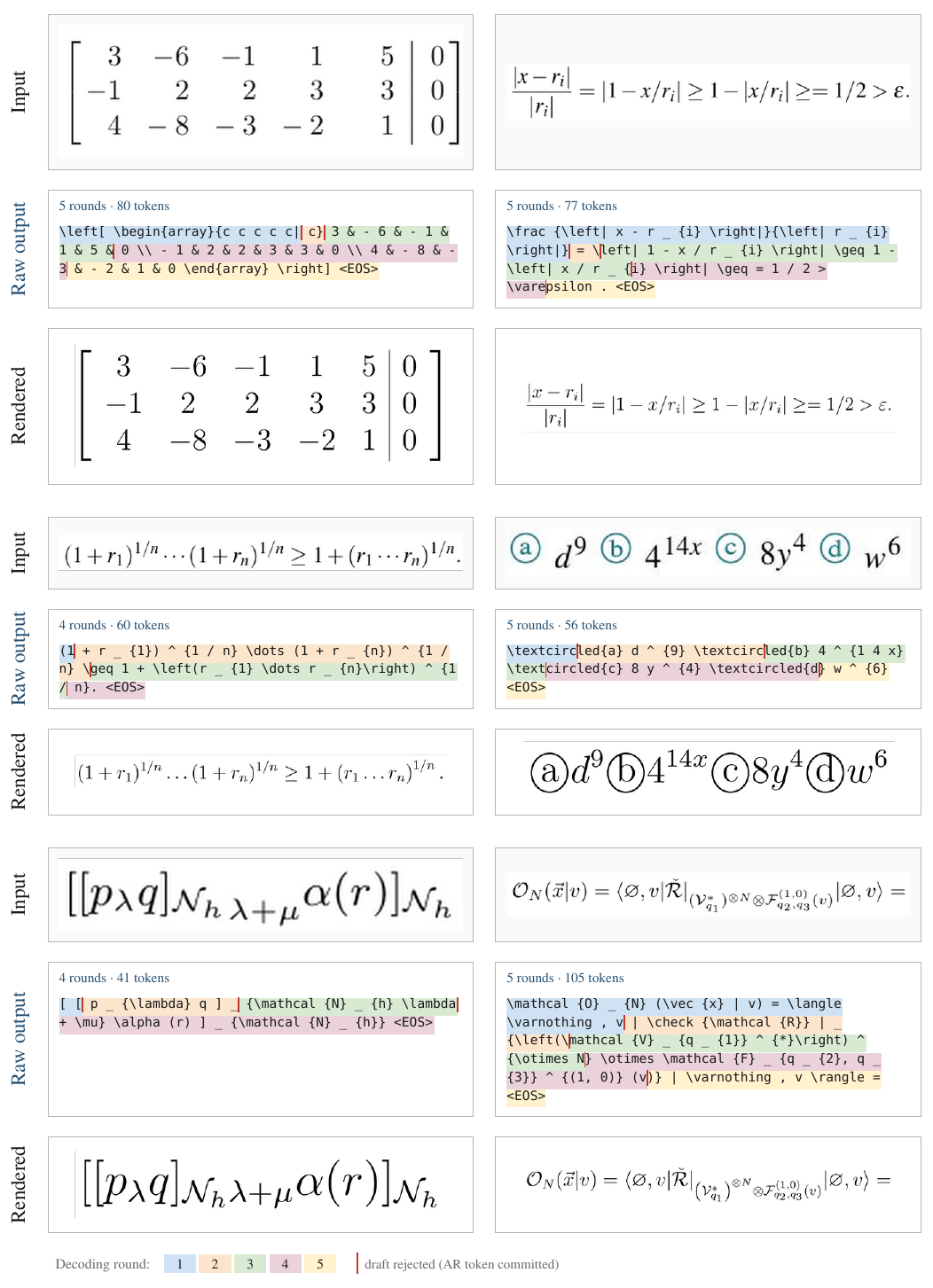}
  \caption{\textbf{Qualitative results on formula regions.} Self-speculative decoding on OmniDocBench.
  Each block shows the input crop, the \LaTeX{} output and its rendering; each color denotes a different
  decoding round, a red tick marks a rejected draft token, and \texttt{<EOS>} is the end-of-sequence token.}
  \label{fig:qual-formula}
\end{figure}

\begin{figure}[p]
  \centering
  \includegraphics[width=0.98\linewidth,height=0.9\textheight,keepaspectratio]{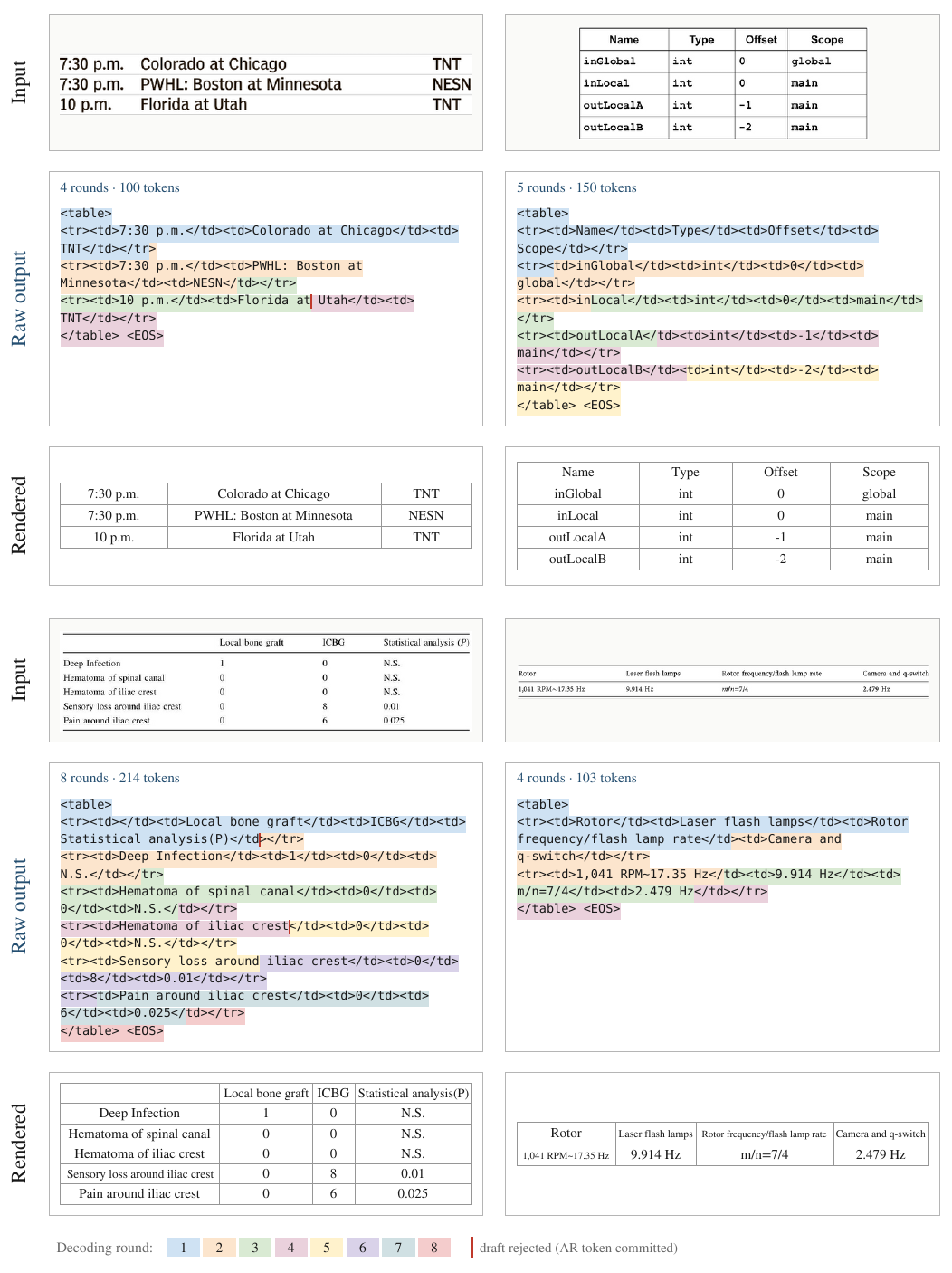}
  \caption{\textbf{Qualitative results on table regions.} Self-speculative decoding on OmniDocBench.
  Each block shows the input crop, the HTML output and its rendering; each color denotes a different
  decoding round, a red tick marks a rejected draft token, and \texttt{<EOS>} is the end-of-sequence token.}
  \label{fig:qual-table}
\end{figure}

\cref{fig:qual-text,fig:qual-formula,fig:qual-table} show self-speculative decoding with the final GRPO
checkpoint on OmniDocBench crops of the three region types. For each crop we show the raw serialized
output with every character shaded by the decoding round that committed it; one round is one
bidirectional draft forward over the masked block followed by one causal verify forward, so the number
of shaded segments is the number of forward pairs that produced the output. A red tick marks a position
where the verifier rejected the draft and committed its own token instead. Every round commits \(a_0\),
the accepted draft prefix and the verifier's own next token, so each round advances by at least two tokens;
\texttt{<EOS>} marks the end-of-sequence token, shaded with the round that emitted it. Each figure carries a legend of the rounds it contains, and each block is headed by its
number of rounds and tokens. For all crops shown, the self-speculative output is byte-identical to
autoregressive greedy decoding of the same weights.

%% file: tables/length_domain_speed.tex
\begin{table}[t!]
\centering
\caption{\textbf{Parallelism and speed by output length.} AR and self-speculative decoding results across output-length ranges. \(n\) is the number of crops in each range. Throughput includes vision encoding, prompt prefill, and decoding; speedup is relative to AR.}
\label{tab:length_domain}
\small
\setlength{\tabcolsep}{5pt}
\begin{tabular}{l r c c c c c}
\toprule
Output length & $n$ & accepted tokens$^{\uparrow}$ & TPF$^{\uparrow}$ & AR tok/s & self-spec tok/s & speedup \\
\midrule
$<128$        & 7{,}257 & 14.8 & 8.4 & 357 & 510  & 1.43$\times$ \\
128--511      & 1{,}496 & 18.1 & 10.1 & 583 & 1{,}176 & 2.02$\times$ \\
512--1023     & 92 & 18.4 & 10.2 & 737 & 2{,}033 & 2.76$\times$ \\
$\geq 1024$  & 77 & \textbf{23.9} & \textbf{12.9} & 761 & \textbf{2{,}805} & \textbf{3.68$\times$} \\
\bottomrule
\end{tabular}
\end{table}

%% file: figures/draft_threshold_sweep.tex
\begin{figure}[t!]
\centering
\definecolor{vizblue}{HTML}{2A78D6}
\definecolor{vizorange}{HTML}{EB6834}
\definecolor{vizgrid}{HTML}{E4E4E0}
\definecolor{vizink}{HTML}{52514E}
\pgfplotsset{
  dsweep/.style={
    width=0.335\linewidth, height=4.4cm, scale only axis,
    xmin=0.44, xmax=1.05,
    xtick={0.5,0.7,0.8,0.9,0.99}, xticklabels={.5,.7,.8,.9,.99},
    xlabel={draft threshold $\tau_d$},
    axis x line*=bottom, axis y line*=left,
    tick align=outside,
    every tick label/.append style={font=\scriptsize, color=vizink},
    label style={font=\small, color=vizink},
    axis line style={vizink, line width=0.5pt},
    ymajorgrids, major grid style={vizgrid, line width=0.35pt},
    clip=false,
  }
}
\subfigure[Accepted draft tokens per round]{%
\begin{tikzpicture}
\begin{axis}[dsweep,
  ylabel={accept.\ $A$\,$\uparrow$},
  ymin=17.6, ymax=26.2, ytick={18,20,22,24,26},
]
\addplot[vizblue, line width=1.1pt, mark=*, mark size=2.2pt,
         mark options={fill=vizblue, draw=white, line width=0.6pt}]
  coordinates {(0.5,20.26) (0.7,22.58) (0.8,23.63) (0.9,24.50) (0.95,24.76) (0.99,25.19)};
\addplot[vizorange, line width=0.9pt, dashed, forget plot]
  coordinates {(0.44,18.96) (1.05,18.96)};
\node[font=\scriptsize, color=vizorange, anchor=south west, xshift=1pt] at (axis cs:0.44,18.96) {one-shot};
\end{axis}
\end{tikzpicture}%
\label{fig:draftsweep:accept}
}\hfill
\subfigure[Tokens per forward]{%
\begin{tikzpicture}
\begin{axis}[dsweep,
  ylabel={TPF\,$\uparrow$},
  ymin=4.0, ymax=11.1, ytick={4,6,8,10},
]
\addplot[vizblue, line width=1.1pt, mark=*, mark size=2.2pt,
         mark options={fill=vizblue, draw=white, line width=0.6pt}]
  coordinates {(0.5,8.64) (0.7,8.20) (0.8,7.81) (0.9,7.13) (0.95,6.31) (0.99,4.86)};
\addplot[vizorange, line width=0.9pt, dashed, forget plot]
  coordinates {(0.44,10.48) (1.05,10.48)};
\node[font=\scriptsize, color=vizorange, anchor=north west, xshift=1pt] at (axis cs:0.44,10.48) {one-shot};
\end{axis}
\end{tikzpicture}%
\label{fig:draftsweep:tpf}
}
\caption{\textbf{Multi-step drafting versus one-shot drafting.} Instead of filling the masked block in a single forward pass, the draft is refined over several denoising steps: at each step the positions whose confidence exceeds the threshold $\tau_d$ are fixed and the rest are re-drafted, and the finished draft is then verified by the causal AR pass exactly as before, so the committed output is identical at every point. (a) A higher threshold spends more steps on the draft and more of its tokens are accepted per round. (b) Every extra step is another forward pass, and the tokens it adds do not pay for it: tokens per forward falls monotonically, and one-shot drafting (dashed) commits the fewest tokens per round yet delivers the highest tokens per forward.}
\label{fig:draftsweep}
\end{figure}

%% file: tables/fullmask_vs_diffusion.tex
\begin{table}[t!]
\centering
\caption{\textbf{All-mask training does not produce a better draft.} The all-mask model commits fewer tokens per forward at every checkpoint, and its accuracy is behind at five of six.}
\label{tab:fullmask}
\small
\setlength{\tabcolsep}{5pt}
\begin{tabular}{l c c c c c c}
\toprule
tokens per forward$^{\uparrow}$ (self-speculative) & 2k & 4k & 5k & 6k & 8k & 10k \\
\midrule
All-mask & 4.04 & 4.82 & 5.54 & 5.94 & 6.45 & 6.66 \\
Uniform \(t\) (ours)             & \textbf{4.64} & \textbf{5.30} & \textbf{6.04} & \textbf{6.00} & \textbf{6.78} & \textbf{6.75} \\
\midrule
\odb{} Overall$^{\uparrow}$ & 2k & 4k & 5k & 6k & 8k & 10k \\
\midrule
All-mask & 94.49 & 94.59 & 94.75 & \textbf{94.66} & 94.76 & 94.86 \\
Uniform \(t\) (ours)             & \textbf{94.71} & \textbf{94.74} & \textbf{94.76} & 94.55 & \textbf{94.92} & \textbf{95.02} \\
\bottomrule
\end{tabular}
\end{table}